\documentclass[letterpaper, 10 pt, conference]{ieeeconf} 

\IEEEoverridecommandlockouts      
\usepackage{times}

\usepackage{multicol}   
\usepackage[bookmarks=true]{hyperref}% This command is only needed if 
\usepackage{amsmath} % assumes amsmath package installed
\usepackage{amssymb}  % assumes amsmath package installed
\usepackage{dsfont}

\usepackage{bm}
\usepackage{amsmath,cleveref} % assumes amsmath package installed
\usepackage{amssymb}  % assumes amsmath package installed
\usepackage{dsfont}
\usepackage{graphicx}
\usepackage{psfrag,graphicx,epsfig}
\usepackage{epstopdf}
\usepackage{xspace}
\usepackage{subcaption}
\usepackage{float}
\usepackage{placeins}
\usepackage{multirow}
\usepackage{pgf,tikz}
\usepackage{nowidow}
\usepackage{lineno}
\usepackage{xcolor}
\usepackage{color}
\usepackage{siunitx}
\usepackage{color}
\usepackage{lineno}
\usepackage{algorithmic}% Needed to meet printer requirements.
\usepackage{gensymb}
\newcommand{\fig}[1]{Fig.~\ref{#1}}
\newcommand{\tab}[1]{Table~\ref{#1}}
\newcommand{\eq}[1]{(\ref{#1})}
\newcommand{\revise}[1]{\textcolor{black}{#1}}
\newcommand{\drake}{\textsc{Drake}}

\usepackage{siunitx}
\usepackage{color}
\usepackage{flushend}
\usepackage{lineno}
\usepackage{gensymb}
\usepackage{algorithm,verbatim}
\usepackage{color-edits}
\addauthor{ms}{magenta}
\addauthor{jw}{green}
\addauthor{tp}{red}
\addauthor{tz}{orange}
\addauthor{xn}{purple}
\addauthor{hz}{brown}
\allowdisplaybreaks
\begin{document}

\title{
% Is Contact Smoothing Better for Planning than Feedback Control?
\LARGE \bf
Is Linear Feedback on Smoothed Dynamics Sufficient for Stabilizing Contact-Rich Plans?
} 
%{for Robust Planning and Control? A Perspective on Whole-Body Manipulation

\author{Yuki Shirai$^{1, 2}$, Tong Zhao$^{2}$, H.J. Terry Suh$^{3}$, Huaijiang Zhu$^{2, 4}$, \\Xinpei Ni$^{2, 5}$,  Jiuguang Wang$^{2}$, Max Simchowitz$^{3}$, and Tao Pang$^{2}$
\thanks{$^1$University of California, Los Angeles, $^2$RAI Institute (formerly Boston Dynamics AI Institute), $^3$MIT CSAIL, $^4$New York University, $^5$Georgia Tech. This work was done during YS, HZ and XN's internship at the RAI Institute. Correspondence to {\tt\small yukishirai4869@g.ucla.edu}.}%
}

\maketitle

%%%%%%%%%%%%%%%%%%%%%%%%%%%%%%%%%%%%%%%%%%%%%%%%%%%%%%%%%%%%%%%%%%%%%%%%%%%%%%%%

\begin{abstract}
% what we want to say:
% Designing LQR is challenging using contact smoothing.
% thanks to robust optimization, LQR performance improves (robust optimization has some challenges though)
%However, contact smoothing has not been exploited well for the design of robust planners and controllers. 
% \hzcomment{The transition here is a bit abrupt and I feel it is not highlighted enough that smooth and differentiable contact dynamics allows us synthesizing LQR gains on the object state.}
% Designing planners and controllers for contact-rich manipulation is extremely challenging because underlying dynamics are non-smooth, resulting in non-convex optimization problems with discontinuous gradients.
% The contact smoothing tool relaxes these limitations and has shown impressive results in open-loop planning for contact-rich manipulation on a fixed model.

Designing planners and controllers for contact-rich manipulation is extremely challenging as contact violates the smoothness conditions that many gradient-based controller synthesis tools assume. 
Contact smoothing approximates a non-smooth system with a smooth one, allowing one to use these synthesis tools more effectively.
% The contact smoothing tool allows us to make a smooth approximation to a non-smooth system, enabling us to use these synthesis tools more effectively. 
% at the cost of introducing small bias.
However, applying classical control synthesis methods to  smoothed contact dynamics remains  relatively under-explored.
This paper analyzes the efficacy of linear controller synthesis using differential simulators based on contact smoothing.  We introduce natural baselines for leveraging contact smoothing to compute (a) open-loop plans robust to uncertain conditions and/or dynamics, and (b) feedback gains to stabilize around open-loop plans. Using robotic bimanual whole-body manipulation as a testbed, we perform extensive empirical experiments on over 300 trajectories and analyze why LQR seems insufficient for stabilizing contact-rich plans.
The summary and the experiments videos can be found \href{https://youtu.be/HLaKi6qbwQg?si=2IucjMBtXp5Sd9Ji}{here}.

\end{abstract}

\IEEEpeerreviewmaketitle
\section{Introduction}\label{sec:intro}

% \tzcomment{Feels like there ought to be more citations in the intro.}
Dexterous manipulation is full of contact-rich interactions, enabling various tasks through complex frictional interactions \cite{mason2018toward, shirai2025hierachical}.
% , when properly exploited, 
% allows us to conduct various tasks through contact with complex frictional interactions . 
\revise{Historically, the non-smooth nature of contact has precluded a range of planning and control methods that
rely on gradients of the dynamics.
% However, the nonsmooth nature of contact dynamics poses many challenges for conventional approaches designed for smooth systems.
% 
Recent advances have utilized \emph{contact smoothing} --- where non-smooth dynamics are replaced by a continuously differentiable proxy --- to great effect as surrogate dynamics models for \emph{planning} through contact \cite{posa2014direct, onol2019contact, dojo, pang2022global}. One may hope, then, that smoothing enables the use  gradient-based \emph{control}.}
% For example, smoothed dynamics models introduce nonphysical phenomena (e.g., force at a distance) which, though mild in their effects on planning, can have major repercussions on the stability of controllers.
% \tzcomment{are there additional reasons why we might, a priori, expect control w/ smoothed models to be hard?}

\begin{figure}[t]
    \begin{subfigure}[t]{0.5\textwidth}
        \centering
        \includegraphics[width=0.99\textwidth]{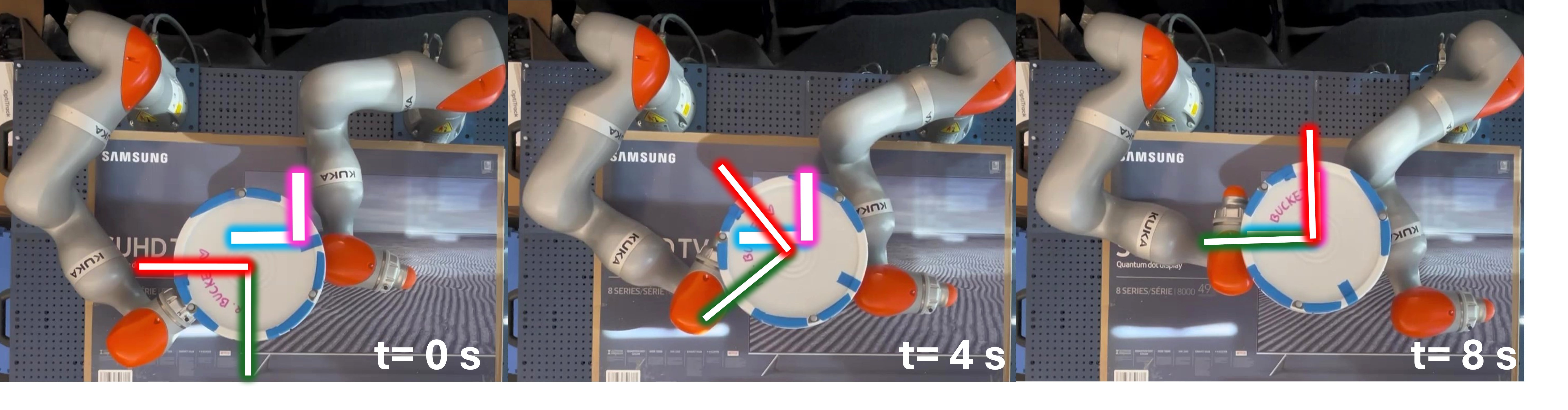}
        \vspace{-1.75em}
        \caption{LQR.}
        \label{fig:fig1a_lqr}
    \end{subfigure}
    \begin{subfigure}[t]{0.5\textwidth}
        \centering
        \includegraphics[width=0.99\textwidth]{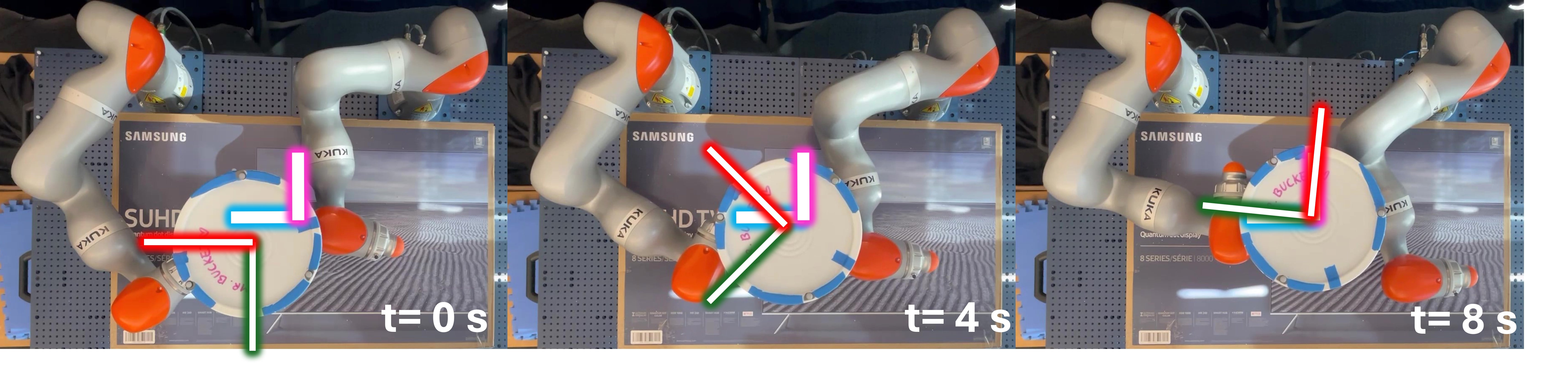}
        \vspace{-1.75em}
        \caption{Open-loop controller.}
        \label{fig:fig1b_open}
    \end{subfigure}
    \caption{
    % We study a method for designing LQR using contact smoothing in the context of bimanual whole-body manipulation.
    % , where two robot arms make contact with an object not only at their end-effectors but also at other parts of the arms. 
    These figures show snapshots of hardware experiments using LQR and open-loop controllers under perturbations to initial conditions of cylinder. 
    The thick and thin lines represent the desired frame at the terminal time step and the current frame of the cylinder, respectively. While LQR outperforms open-loop in this example, a more comprehensive evaluation shows that LQR generally performs poorly. 
    The hardware experiment videos can be found \href{https://youtu.be/HLaKi6qbwQg?si=2IucjMBtXp5Sd9Ji}{here}.
    % We discuss when and why LQR works/does not work. 
    % \textcolor{orange}{[tz: Suggest only displaying x axis to show rotation angle; having both x and y axes present makes the diagram feel cluttered to me. In the contact segment distribution plots, we use blue for the object's start configuration and red for the object's end configuration. Also suggest labeling start and goal in text on the image]}
    }
    \label{fig:whole-body}
\end{figure}

This work suggests that the above hope  may face significant obstacles. We \textbf{(1)} introduce 
% of using smoothed dynamics models for 
LQR control for contact-manipulation via contact smoothing. 
% We investigate the performance of LQR using contact smoothing.
Furthermore, we \textbf{(2)} present and analyze robust trajectory optimization, hoping that the generated trajectories are robust to the model errors accumulated by using a surrogate dynamics model for control, and thus more amenable to LQR. Then, we \textbf{(3)} extensively evalute the performance of these methods, both in simulation and in hardware, on a bimanual whole-body manipulation as shown in \fig{fig:whole-body}. In short, we find:
\begin{quote}
    \emph{Despite its efficacy in planning through contact, dynamical smoothing alone is unsatisfactory as a means to obtaining linear control policies.}
\end{quote}
Finally, we \textbf{(4)} identify the key factors leading to the inadequacies of linear control; namely, the \emph{unilaterality} of contact,  and the tendency of controllers to ``push and pull'' unless the dynamics are only very-slightly smoothed.

\section{Related Work}\label{sec:related_works}
%In this section, we go over some works which are closely related to the work presented in this paper. 

\subsection{Planning through Contact}
% for planning
% it is very difficult due to non-smooth, challenging
% for control
% online execusion. also estimator is difficult
% say in this work, we 
%Planning through contact is difficult because the motion planners need to reason non-smooth dynamical systems. Yet, there have been a number of impressive works in this area, especially focusing on locomotion and manipulation problems. 
One approach 
% \hzcomment{nit: is time stepping a motion planning approach itself or rather a component of the approach?} 
to motion-planning through contact enforces contact dynamics by including linear complementarity constraints \cite{stewart1996implicit, posa2014direct, carius2019trajectory, onol2019contact, patel2019contact, Jeon2022Online}. 
Another popular method to address contacts is based on Mixed-Integer Programming
% \hzcomment{I would just say ``based on mixed-integer programing'' here. The phrase integer constraints is ambiguous (is it integer-valued or taking integer-valued variables or both?)}, leading to Mixed-Integer Programming 
(MIP) \cite{ding2020kinodynamic, cauligi2020learning, marcucci2020warm, shirai2022simultaneous, Wensing2024Optimization}, where discrete variables encode contact \textcolor{black}{modes}. However, due to the inherent non-smoothness of contact \textcolor{black}{dynamics}, both methods suffer from poor scalability as the number of contact modes increases.

%However, in general, solving LCP and MIP is computationally challenging because they do not scale as the number of decision variables and the constraints increase. 
To tackle this,
% the poor scalability in planning methods due to the non-smoothness of contact dynamics, 
contact smoothing has been proposed, replacing exact dynamics models with surrogate models that obey second-order smoothness \cite{posa2014direct, dojo, suh2022bundled, pang2022global}.
This approach introduces a ``force-at-a-distance'' effect where gradients through the dynamics convey information about nearby contacts. Since gradients of the dynamics become continuous, gradient-based optimizers are able to find solutions more efficiently.
% 
% One advantage is that 
%With this motivation, some works consider solving smooth approximations of the contact dynamics because the optimization solver is able to get the informative gradients of the system. 
\textcolor{black}{Our} work employs a log-barrier smoothing scheme \cite{dojo, pang2022global}, which is more efficient compared to stochastic smoothing schemes \cite{suh2022bundled, diffsim}.
% \mscomment{@Pang maybe take a pass here}
% Using this contact smoothing in optimization for planners, we can efficiently design various contact-rich trajectories.
%\, which stochastically smooths the non-smooth contact dynamics. Another advantage of using randomized smoothing is that it enables the users to use the traditional robust optimization techniques, which are often limited to continuous dynamical systems. 

% Inspired by this motivation, in this paper, we employ the randomized smoothing to compute the gradients through contact. 

\subsection{Control through Contact}
% \revise{we want to say we don't do potentially computationally expensive MPC. Instead, we do minimum controller which still uses the underlying contact structure, i.e., LQR. We don't do RL as well. }
% force controller like alp one. 
% estimator
Prior approaches to control through contact reason about contact modes  \cite{hogan2018reactive, aydinoglu2022real, moura2022non, shirai2023covariance}. 
% \hzcomment{I don't think Simon's paper explicitly reasons about contact modes?}
% 
% For example, the work presented in \cite{hogan2018reactive} proposes to employ a  supervised learning module for selecting contact mode schedule, and thus the resulting optimization problems become a convex quadratic program from MIP. 
While these approaches can run in real time for simple systems with only a handful of contact modes, they have yet to scale to high-dimensional systems, multiple objects per scene, or contacts with objects of complex and/or irregular shapes due to the expensive computation.
% Model Predictive Control (MPC) might not work once there are many constraints 

In contrast, contact-smoothing offers a way to directly apply \emph{smooth} control methods while being less constrained by the non-smoothness of contact dynamics, at the cost of introducing some smoothing bias \cite{suh2022bundled}. 
In this work, we study LQR, which uses the underlying problem structure provided by contact smoothing, enabling its application to high-dimensional bimanual whole-body manipulation tasks.

\subsection{Robust Planning through Contact}\label{sec:robust_planning_related_work}
%Recently, there has been some work for designing robust manipulation planners and controllers using robust and stochastic optimization. 
% \cite{hammoud2021impedance}.
% Past work on robust planning through contact falls roughly into one of two categories. In \emph{domain randomization}, one attempts stochastic optimization over a fixed distribution of dynamics models \cite{ shirai2023covariance, shirai2024chance}. \emph{Worst-case optimization}, the more conservative approach, optimizes for performance on a worst-case model in a given uncertainty set  \cite{dai2016planning, orsolino2018application, shirai2022robust, shirai2024robust}. 
% While those works show remarkable results, they design robust trajectories against parametric uncertainties in the system dynamics such as mass and coefficients of friction and they do not discuss uncertainty in the shape of objects (e.g., the radius of a cylinder).
% Considering shape uncertainty is quite important for generalized manipulation to avoid unexpected contact events.
% However, designing robust trajectories against shape uncertainty is difficult since it is not clear how to consider shape uncertainty in the system dynamics. 
% In this work, we incorporate domain randomization with an emphasis on uncertainty in shape. Using contact smoothing, our smoothed contact dynamics is able to consider smoothed collision dynamics, enabling robust optimization to consider the shape uncertainty.
% Also, we hope that using our technique enables LQR to track reference trajectories more easily. 

Previous work on robust planning through contact (e.g., \cite{mordatch2015ensemble}) generally falls into two approaches: \emph{domain randomization}, which involves stochastic optimization over a fixed distribution of dynamics models \cite{shirai2023covariance, shirai2024chance}, and \emph{worst-case optimization}, which focuses on performance under worst-case scenarios within an uncertainty set \cite{dai2016planning, orsolino2018application, shirai2022robust, shirai2024robust}. These methods effectively handle parametric uncertainties in the system dynamics, such as mass and friction, but do not discuss shape uncertainty (e.g., the radius of a cylinder). Shape uncertainty is critical for generalized manipulation to avoid unexpected contact events, yet integrating it into system dynamics remains challenging.

In this work, we incorporate domain randomization with an emphasis on shape uncertainty for primitive objects.
Using contact smoothing, our method incorporates smoothed collision dynamics, enabling robust optimization to consider shape uncertainty. We also hope that using our technique enables LQR to track reference trajectories more easily.

% and thus robust optimization is able to consider the shape uncertainty.
% \hzcomment{``in the system dynamics''}.
% \hzcomment{I feel like the flow is better if we put the sentence ``In this work...'' after ``...take into account the shape uncerstainty''. This way we immediately highlight that contact smoothing made it possible.}

% Although we only focus on domain randomization in this work, our algorithm can consider worst-case optimization as well through contact smoothing. 

%Another popular direction is robust optimization where the optimization finds the robust trajectory given the uncertain set, not the distribution. Since it is challenging to assess the distribution of the states for contact-rich systems, robust optimization techniques are more commonly used in contact-rich tasks \cite{dai2016planning, del2018zero, orsolino2018application, shirai2022robust}. 
% While these works show impressive results by solving min-max problems, in this work we propose to utilize the excess risk \cite{bubeck2012regret, hazan2016introduction, simchowitz2021statistical}.
\section{Problem Statement}\label{sec:problem_statement}
%We now give a mathematical description of our problem setting, which pertains to 

%The formulations that follow consider a quasi-dynamic dynamic 

% In this section, we describe our problem statement. 
We focus on the manipulation of rigid objects. Throughout, we focus on bimanual tabletop manipulation, though in principle our approach extends beyond this regime. 

% \begin{comment}
% \begin{enumerate}
%     \item The object is rigid.
%     \item We consider quasi-dynamic dynamics of the object and the robot. 
%     \item We consider tabletop manipulation. \hzcomment{Is this necessary? In principle this approach works in other scenarios?}
%     % \item 
% \end{enumerate}
% \end{comment}

\subsection{Quasi-Dynamic Dynamics Model}

% \tpcomment{I feel sub-sections A, B and C can be moved into the appendix? All we need in later sections is $x_{t+1} = f(x_t, u_t; p)$, and the internal workings of the contact dynamics doesn't seem super relevant.}

We generate plans and linear feedback gains by considering a quasi-dynamic model of a robot manipulating a single rigid object \cite{anitescu2006optimization}. We consider robots with $n_{\mathrm{a}}$ actuated Degrees of Freedom (DoFs) and the objects with $n_{\mathrm{u}}$ unactuated DoFs. We denote the configurations of the object and the robot as $\mathbf{q}^{\mathrm{u}} \in \mathbb{R}^{n_{\mathrm{u}}}$ and $\mathbf{q}^{\mathrm{a}} \in \mathbb{R}^{n_{\mathrm{a}}}$, respectively. 
In quasi-dynamics models, we assume velocities are small, and thus system states are the concatenation of both configurations $\mathbf{x}:=\mathbf{q}:=\left[{\mathbf{q}^{\mathrm{u}}}^\top, {\mathbf{q}^{\mathrm{a}}}^\top\right]^\top\in \mathbb{R}^{n_{\mathrm{x}}}$, where $n_{\mathrm{x}} = n_{\mathrm{u}} + n_{\mathrm{a}}$. \footnote{\revise{Superscript $\mathrm  u$ stands for unactuated, and $\mathrm a$ for actuated.}}
We denote the change in system configuration from the current time step to the next time step as  $\boldsymbol{\delta}\mathbf{q}:=\left[\boldsymbol{\delta}{\mathbf{q}^{\mathrm{u}}}^\top, \boldsymbol{\delta}{\mathbf{q}^{\mathrm{a}}}^\top\right]^\top$. 
We denote the control input as  $\mathbf{u} \in \mathbb{R}^{n_{\mathrm{a}}}$, defined as the commanded positions of the robot's joints. 
\revise{We consider the linear feedback law $\mathbf{u} = \mathbf{v} + \mathbf{K}\mathbf{\Delta x}$ where $\mathbf{v} \in \mathbb{R}^{n_{\mathrm{a}}}$ is the feedforward gain from trajectory optimization, $\mathbf{K} \in \mathbb{R}^{n_{\mathrm{a}}\times n_{\mathrm{x}}}$ is the feedback gain computed by LQR, and $\mathbf{\Delta x}$ is the state error, or deviation from nomial trajectory.}

\subsection{Quasi-Dynamic Dynamics Model with Contact Smoothing}\label{sec:Derivatives of Forward Dynamics}
% \tpcomment{I recommend getting rid of most of this subsection except for equation (2) and (3). We can refer the readers to the global planning paper for definitions of the jacobians and the log barrier.}

For computing $\boldsymbol{\delta}\mathbf{q}$, we consider a log-barrier-smoothed formulation of quasi-dynamic dynamics from Eq. (34) in \cite{pang2022global}. We denote $\kappa$-smoothed forward dynamic map as
\begin{equation}
    \mathbf{q}_{t+1} = f_{\kappa}\left(\mathbf{q}_t, \mathbf{u}_t\right)
    \label{eq:smoothed_forward_dynamics}
\end{equation}
where $\kappa > 0$ is the user-defined barrier parameter. Smaller $\kappa$ corresponds to greater contact smoothing, and more ``force at a distance'' \cite{pang2022global}.
Recalling our notation $\mathbf{x} = \mathbf{q}$, the above can also be written as $\mathbf{x}_{t+1} = f_{\kappa}\left(\mathbf{x}_t, \mathbf{u}_t\right)$. When $\kappa$ does not change, we will simply write $f$ for $f_{\kappa}$. 

% \textbf{Parametrizing the dynamics.} We allow for \emph{uncertainty over the dynamics} induced by object uncertainty.  These can be parameterized by parameter $\mathbf{p} \in \mathbb{R}^{p}$ which enters into the dynamics through the dependencies $\mathbf{M}_{\mathrm{u}} = \mathbf{M}_{\mathrm{u}} (\mathbf{q};\mathbf{p})$. and $\mathbf{M}_{\mathrm{u}} = \mathbf{M}_{\mathrm{u}} (\mathbf{q};\mathbf{p})$ and $\mathbf{J}_{\mathrm{u},i}(\mathbf{q},\mathbf{p})$. \mscomment{make consistent.}
% \textbf{Derivatives of Forward Dynamics}\label{sec:Derivatives of Forward Dynamics}
Both our gradient descent-based trajectory optimizer and our synthesis of linear feedback rely on differentiation of the smoothed dynamics. Given \eq{eq:smoothed_forward_dynamics}, we define derivatives: 
\begin{equation}
\mathbf{A}_{\kappa}=\frac{\partial f_{\kappa}}{\partial \mathbf{q}}, \ \quad \mathbf{B}=\frac{\partial f_{\kappa}}{\partial \mathbf{u}}.
\label{derivatives of forward dynamics}
\end{equation}
See \cite{pang2022global} for the complete derivation of \eq{derivatives of forward dynamics}.
% We describe how these can be computed via the implicit function theorem in \Cref{sssec:comp:implicit_fnc_thm}.
Again, we use the shorthand $\mathbf{A}=\mathbf{A}_\kappa$ and $\mathbf{B}=\mathbf{B}_\kappa$.

\subsection{Parametrized Quasi-Dynamic Dynamics Model}\label{sec:Parametrized}

% \tzcomment{Should argue that our parameterization of the uncertainty here is motivated by LQR's missing of contacts. Not sure how to make this argument though.}
As described in Sec~\ref{sec:robust_planning_related_work}, 
we consider \emph{uncertainty over the dynamics} induced by object uncertainty. 
% Thus, we hope that 
% 
These can be parameterized by a parameter $\mathbf{p} \in U \subseteq \mathbb{R}^{n_{\mathrm{p}}}$ which enters into $f_\kappa$ in \eq{eq:smoothed_forward_dynamics}.
% which enters into the dynamics through the matrix $\mathbf{Q}$, linear term $\mathbf{b}$, and contact Jacobians  $\mathbf{J}_{\mathrm{t}_i}$, $\mathbf{J}_{\mathrm{n}_i}$ in (34) in \cite{pang2022global} such as $\mathbf{M}_{\mathrm{u}} = \mathbf{M}_{\mathrm{u}} (\mathbf{q};\mathbf{p})$, $\mathbf{J}_{\mathrm{u},i}(\mathbf{q};\mathbf{p})$, and $\mathcal{K}_i^{\star}(\mathbf{p})$. 
% These dependencies are explained in detail in \mscomment{todo??}. 
To lighten notation, we also let the parameter $\mathbf{p}$ encode an initial condition, $\mathbf{x}_{\mathrm{init}}(\mathbf{p})$.

%dependencies $\mathbf{M}_{\mathrm{u}} = \mathbf{M}_{\mathrm{u}} (\mathbf{q};\mathbf{p})$, $\mathbf{J}_{\mathrm{u},i}(\mathbf{q};\mathbf{p})$, and $\mathcal{K}_i^{\star}(\mathbf{p})$\hzcomment{These quantities are not defined.}, where $U$ represents the uncertain set of $\mathbf{p}$. In this work, we focus on uncertain shape parameters such as the length and width of the object. The induced dynamics are then $f_{\kappa}(\mathbf{x},\mathbf{u};\mathbf{p})$, again suppressing dependence on $\kappa$ when clear.
% \mscomment{make consistent.}

% We denote uncertain parameters as $\mathbf{p} \in {U}$ where $\mathbf{p}$ is sampled by the uncertain set ${U} \in \mathbb{R}^{n_\mathrm{p}}$. 
% In this project, we explicitly consider the following uncertain parameters.
% \begin{itemize}
%     \item Inertia
%     \item Coefficient of friction
%     \item Shape
% \end{itemize}
% Based on my experience, I feel that shape uncertainty is the most challenging since the robots might lose contact, which is catastrophic in 3D scenarios. We will first work on a similar shape, meaning that just the size of the shape can be different.
\section{Contact-Implicit Trajectory Optimization}\label{sec:robust_optimization}

In this section, we present both single-parameter and multi-parameter trajectory optimization baselines. 

\subsection{Single-Parameter Trajectory Optimization (SP-TrajOPT)}
We formulate our optimal \revise{planning} problem as follows. 
\begin{subequations}
\begin{flalign}
\min _{\{\mathbf{x}_t\}_{t=1}^T, \{\mathbf{v}_t\}_{t=0}^{T-1}} &\; 
J\left(\{\mathbf{x}_t\}_{t=1}^T, \{\mathbf{v}_t\}_{t=0}^{T-1};\mathbf{p}\right)
\\
\text{s.t.} &\; \mathbf{x}_{t+1} = f\left(\mathbf{x}_t, \mathbf{v}_t; \mathbf{p}\right), \forall t \in \mathcal{T}\\
&\; g\left(\mathbf{x}_t, \mathbf{v}_t; \mathbf{p}\right) \leq \mathbf{0}, \forall t \in \mathcal{T} \label{eq:joint torque}\\
&\; \mathbf{x}_t \in \mathcal{X}, \mathbf{v}_t \in \mathcal{V}, \forall t \in \mathcal{T} \\
 &\; \mathbf{x}_0 = \mathbf{x}_{\mathrm{init}} \left(\mathbf{p}\right),
\label{robust_TO_initial_bound}
\end{flalign}
\label{equation_single_TO}
\end{subequations}
where $\mathcal{T} := \{0, \ldots, T - 1\}$,  $J$ is the trajectory-wise cost function, $f = f_{\kappa}$ is the contact dynamics of the system (see \eq{eq:smoothed_forward_dynamics}). $\mathbf{v}_t$ is the control input at time step $t$. \eq{eq:joint torque} is used to impose inequality constraints involving $\mathbf{x}_t$ and $\mathbf{v}_t$ such as joint torque constraints and non-collision constraints. $\mathcal{X}$ and $\mathcal{V}$ represent a convex polytope, consisting of a finite number of linear inequality constraints for bounding the decision variables. In \eq{robust_TO_initial_bound}, the initial condition encoded by $\mathbf{p}^{i}$ determines the initial state $\mathbf{x}_{0}$
% $\mathbf{x}_{0} = \mathbf{x}_{\mathrm{init}}^{i} \left(\mathbf{p}^{i}\right)$ 
(see \Cref{sec:Parametrized}).

% \revise{I noticed that TrayOPT is dramatically simplified. Can we go back to the original presentation? like showing SP and then showing MP and thus it is natural to show the transition from the basic trajopt to trajeopt considering robustness?} % TZ: resolved over slack

\subsection{Multi-Parameter Trajectory Optimization (MP-TrajOPT)}
\label{Sec:particle_robust_TO}
% \revise{WIP}
% In order to generate trajectories robust to uncertainty in $\mathbf{p}$, we formulate a robust trajectory optimization problem over realizations of the uncertainty. 
An alternative to \eq{equation_single_TO} is a robust formulation over multiple parameters.
We focus on the simplest approach: optimizing average performance on $N$ realizations $(\mathbf{p}^i)_{1 \le i \le N}$, which for simplicity are manually chosen, inspired by \cite{Blackmore2010probabilistic, shirai2023covariance}, using Sample Average Approximation (SAA) \cite{luedtke2008saa}. For each $\mathbf{p}^i$, we optimize over a corresponding trajectories $(\mathbf{x}^i)_{1 \le i \le N}$. 
% Thus, our trajectory optimization problem is given by:
% 
% We formulate our optimal control problem as follows. 
\begin{subequations}
\begin{flalign}
\min _{\mathbf{X}, \{\mathbf{v}_t\}_{t=0}^{T-1}} &\; 
\frac{1}{N}
\sum_{i =1}^N J\left(\{\mathbf{x}_t^i\}_{t=1}^T, \{\mathbf{v}_t\}_{t=0}^{T-1};\mathbf{p}^i\right) \label{eq:cost}
\\
\text{s.t.} &\; \mathbf{x}^i_{t+1} = f_{\kappa}\left(\mathbf{x}^i_t, \mathbf{v}_t; \mathbf{p}^i\right), \forall t \in \mathcal{T}, \forall i \in \mathcal{I} \label{eq:particle-average-dynamics}\\
&\; g\left(\mathbf{x}^i_t, \mathbf{v}_t; \mathbf{p}^i\right) \leq \mathbf{0}, \forall t \in \mathcal{T}, \forall i \in \mathcal{I} \label{eq:joint-torque}\\
&\; \mathbf{x}^i_t \in \mathcal{X}, \mathbf{v}_t \in \mathcal{V}, \forall t \in \mathcal{T}, \forall i \in \mathcal{I} \label{eq:particle-average-bounds} \\
 &\; \mathbf{x}^i_0 = \mathbf{x}_{\mathrm{init}} \left(\mathbf{p}^i\right), \forall i \in \mathcal{I} \label{eq:particle-average-initial-bound}
\end{flalign}
\label{equation_average_cost_particle}
\end{subequations}
% 
% We perform the optimization over
where $\mathbf{X}:= \{\mathbf{x}^i_t , \forall t \in \mathcal{T}, \forall i \in \mathcal{I}\}$, $\mathcal{I} := \{0, \ldots, N - 1\}$. 
% and $\{\mathbf{v}_t , \forall t \in \mathcal{T}\}$. 
We emphasize that we do not have superscript $i$ on $\mathbf{v}_t$ because our objective is to design a single plan that succeeds on average over the $N$ parameters. 
% When $N = 1$, we refer to the procedure as SP-TrajOPT (Single-Parameter Trajectory Optimization) as it only considers a single realization of $\mathbf{p}$. For $N > 1$, we refer to the procedure as MP-TrajOPT (Multi-Parameter Trajectory Optimization). 
Note that \eq{equation_average_cost_particle} with
$N=1$ specializes to \eq{equation_single_TO}. We call this SP-TrajOPT (Single-
Parameter Trajectory Optimization) as it only considers a
single parameter. For $N > 1$, we refer to the procedure as
MP-TrajOPT (Multi-Parameter Trajectory Optimization).

\section{Feedback Synthesis via LQR}\label{sec:lqr}
In this section, we present an approach to feedback gain synthesis by solving an LQR problem through linearizations of the contact smoothed dynamics.
% We then present a natural extension to LQR through \emph{keypoint} measurements of the object, which (a) do not require full state observation and (b) can encode shape information via the observation function. 
% In what follows, we use capital $\Delta$ to denote linearized error dynamics, in distinction to lowercase $\delta$ used in \Cref{sec:problem_statement} when defining actual (nominal) system dynamics. 

% \subsection{Time-Varying LQR using Contact Smoothing}\label{sec:tv-lqr}

Using \eq{derivatives of forward dynamics}, we can compute LQR feedback gains $\mathbf{K}$.
% As our system dynamics is nonlinear, 
The objective of using LQR is to design a controller that can locally stabilize the system. In this work, we consider the following optimal control problem given  $\mathbf{x}_{t}$ and $\mathbf{v}_{t}, t\in \mathcal{T}$. 
% which can be computed using \eq{equation_average_cost_particle}.
\begin{subequations}
\begin{flalign} \min_{\{\boldsymbol{\Delta} \mathbf{v}_{t}\}_{t=0}^{T-1}}
&\;\sum_{t = 1}^{T} \left\|\boldsymbol{\Delta} \mathbf{x}_{t} \right\|_{\mathbf{Q}_t}^2+\sum_{t = 0}^{T-1}\left\|\boldsymbol{\Delta} \mathbf{v}_{t} \right\|_{\mathbf{R}_t}^2
% + \left\|\boldsymbol{\Delta} \mathbf{x}_{T} \right\|_{\mathbf{Q}_T}^2
\\
\text{s.t.} &\; \boldsymbol{\Delta}\mathbf{x}_{t+1} = \mathbf{A}_{t}\boldsymbol{\Delta}\mathbf{x}_{t} + \mathbf{B}_{t}\boldsymbol{\Delta}\mathbf{v}_{t}, t \in \mathcal{T}\\
&\; \boldsymbol{\Delta}\mathbf{x}_{0} = \hat{\mathbf{x}} -\mathbf{x}_{0} 
\end{flalign}
\label{eq:lqr}
\end{subequations}
where $\mathbf{Q}_t = \mathbf{Q}_t^\top$ 
% and 
% $\mathbf{Q}_T = \mathbf{Q}_T^\top$
% \hzcomment{No need to write for $\mathbf{Q}_T$ separately since it is just $\mathbf{Q}_t, t=T$ }
is positive semidefinite and $\mathbf{R}_t = \mathbf{R}_t^\top$ is positive definite at $t$. $\hat{\mathbf{x}}$ is the measurement of states at $t = 0$.
$\mathbf{A}_{t}$ and $\mathbf{B}_{t}$ are error dynamics, which are obtained by linearizing true nonlinear contact dynamics of the system $f$ around $\mathbf{x}_{t}$ and $\mathbf{v}_{t}, t\in \mathcal{T}$ in accordance with Sec.\ref{sec:Derivatives of Forward Dynamics}. 
% (again, supressing dependence on smoothing parameter $\kappa$). That is, the error dynamics arise from the implicit differentiation of the smoothed contact dynamics given in \eq{eq:smoothed_forward_dynamics}.
% {\ref{eq:smoothed_socp}.} 
%We can compute $\mathbf{A}_{t}$ and $\mathbf{B}_{t}$ using contact smoothing as we explain in Section \ref{sec:Derivatives of Forward Dynamics}.
% 
\revise{It is worth noting again that $\mathbf{A}_{t}$ and $\mathbf{B}_{t}$ convey local contact information of the system dynamics.}
\revise{We use Riccati recursion \cite{anderson2007optimal} to compute $\mathbf{K}$ for \eq{eq:lqr}.}
% Since the optimal controller for \eq{eq:lqr} is given by the solution to the Riccati recursion \cite{anderson2007optimal}: 
% \begin{subequations}
% \begin{flalign} 
% &\; \boldsymbol{\Delta}\mathbf{v}_{t} = \mathbf{K}_{t} \boldsymbol{\Delta}\mathbf{x}_{t}
% \\
% \text{where} &\;  \mathbf{K}_{t} = -\left(\mathbf{R}_{t}+\mathbf{B}_{t}^\top \mathbf{P}_{t+1}\mathbf{B}_{t}
% \right)^{-1} \mathbf{B}_{t}^\top \mathbf{P}_{t+1} \mathbf{A}_{t} \\
% &\; \mathbf{P}_{t}:=  \mathbf{Q}_{t} + \mathbf{A}_{t}^\top  \nonumber \\&\;\left(\mathbf{P}_{t+1} - \mathbf{P}_{t+1} \mathbf{B}_{t} \left(\mathbf{R}_{t} + \mathbf{B}_{t}^\top \mathbf{P}_{t+1} \mathbf{B}_{t}\right)^{-1} \mathbf{B}_{t}^\top \mathbf{P}_{t+1} \right) \mathbf{A}_{t}\\
% &\; \mathbf{P}_{T} = \mathbf{Q}_{T}
% \end{flalign}
% \label{eq:feedback_controller_lqr}
% \end{subequations}
% 
 % In this work, we compute feedback gain $\mathbf{K}_t, t \in \mathcal{T}$ by providing $\mathbf{A}_{\kappa, t}$ and $\mathbf{B}_{\kappa, t}$ with parameters (i.e., $\mathbf{Q}_t, \mathbf{R}_t, \mathbf{P}_T$). Therefore, we design a local linear feedback controller which takes into consideration the underlying contact dynamics.
% 
% 
% We borrow advantages and disadvantages from LQR.
All of the computation happens offline. Hence, there is no expensive computation online unlike other methods (e.g., model predictive control).

\section{Results}\label{sec:results}

In this section, we demonstrate our proposed controller under various uncertainties such as perturbations of initial conditions and shape variations for a cylinder. Through this section, we answer the following questions. 
\begin{enumerate}
    % \item How much does robust optimization robustify manipulation?
    \item How well do TrajOPT and LQR work?
    \item To what extent does MP-TrajOPT improve the performance of LQR?
    \item Under what circumstances do our proposed controllers succeed or fail? 
    % \item How much robustness is beneficial for controllers to stabilize manipulation?
    % \item How well can our controller perform in hardware experiments under unexpected contact events?
\end{enumerate}
% Through the experiments, we mainly discuss the tracking errors of controllers against perturbations to initial conditions and the radius of the cylinder, $r >0$.

% \revise{Say the spec of the computer.}
% Through the experiments, we mainly discuss tracking errors of trajectory optimization with / without LQR on top of generated trajectories from RRT. 
% \revise{how about this?}
% \hzcomment{The opening of this section needs some rewriting to be more clear and better motivated. We should also preview it here that we are testing robust trajectory optimization and LQR on top of RRT-generated plans.}

 \begin{figure}[t]
    \centering
    \includegraphics[width=0.45\textwidth]{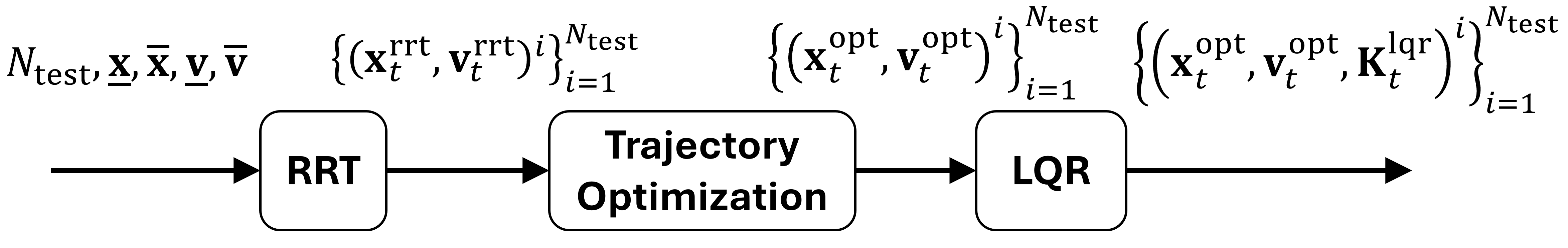}
     \caption{Illustration of our pipeline for generating dataset. We first provide RRT with the number of reference trajectories, $N_\text{test}$, and lower- and upper-bounds of $\mathbf{x}$ and $\mathbf{v}$, denoted as  $\underline{[\cdot]}$ and $\bar{[\cdot]}$, respectively. Next, RRT samples initial and desired terminal states of an object and robots and plans $N_\text{test}$ different trajectories $\{\left(\mathbf{x}^{\text{rrt}}_t, \mathbf{v}^{\text{rrt}}_t\right)^i\}_{i=1}^{N_\text{test}}$. Then, our trajectory optimizers compute $N_\text{test}$ different trajectories of the object and the robots, $\{\left(\mathbf{x}^{\text{opt}}_t, \mathbf{v}^{\text{opt}}_t\right)^i\}_{i=1}^{N_\text{test}}$ using $\{\left(\mathbf{x}^{\text{rrt}}_t, \mathbf{v}^{\text{rrt}}_t\right)^i\}_{i=1}^{N_\text{test}}$ as warm-start. Finally, given $\{\left(\mathbf{x}^{\text{opt}}_t, \mathbf{v}^{\text{opt}}_t\right)^i\}_{i=1}^{N_\text{test}}$, LQR module computes $\{\left(\mathbf{K}_t^\text{lqr}\right)^i\}_{i=1}^{N_\text{test}}$ locally.
     % We use the parameter (\texttt{a}) in \tab{table:parameter_objects} for RRT, SP-TrajOPT, and LQR, and the full set (\texttt{a}), (\texttt{b}), and (\texttt{c}) for MP-TrajOPT
     % For RRT, SP-TrajOPT, and LQR, we use the parameter (\texttt{a}) in \tab{table:parameter_objects} and for MP-TrajOPT, we use all (\texttt{a}), (\texttt{b}), and (\texttt{c}).
     }
\label{fig:data_pipeline}
\end{figure}

\begin{table}[b]
    \caption{Parameters of cylinders.}
    \centering
    \begin{tabular}{c|c|c|c}
    &  mass [kg]  & friction coefficient & radius, length [m]\\
         \hline\hline 
       (\texttt{a}) & 1.0 & 0.5 & 0.14, 0.5 \\
       \hline
       (\texttt{b}) & 1.0 & 0.5 & 0.13, 0.5 \\
              \hline
       (\texttt{c}) & 1.0 & 0.5 & 0.15, 0.5 \\
       \hline 
      %  \hline
      % Box & mass [kg]  & friction coefficient & width, depth, height [m] \\
      % \hline \hline
      %        (a) & 1.0 & 0.5 & 0.45, 0.45, 0.6 \\
      % \hline
      %              (b) & 1.0 & 0.5 & 0.44, 0.45, 0.6 \\
      % \hline
      %                    (c) & 1.0 & 0.5 & 0.46, 0.46, 0.6 \\
      % \hline
    \end{tabular}
    \label{table:parameter_objects}
\end{table}
% Throughout, we consider a cylinder and a box as our two objects of manipulation. 
% Plans and \drake{} simulations are conducted assuming uniform mass density. 
\subsection{Experimental Setup}
Three sets of parameters 
% \tpcomment{Consider using a different font, e.g. \texttt{a}?},
 are in \tab{table:parameter_objects}. For SP-TrajOPT, we use the nominal ``(\texttt{a})'' shape in that table; for MP-TrajOPT we use all $N=3$ ``(\texttt{a})'', ``(\texttt{b})'' and ``(\texttt{c})'' shapes. 
% 
% \subsubsection{Data Generation Pipeline}\label{sec:data_generation_pipeline} 
\begin{figure}[t]
    \centering
    \includegraphics[width=0.4\textwidth]{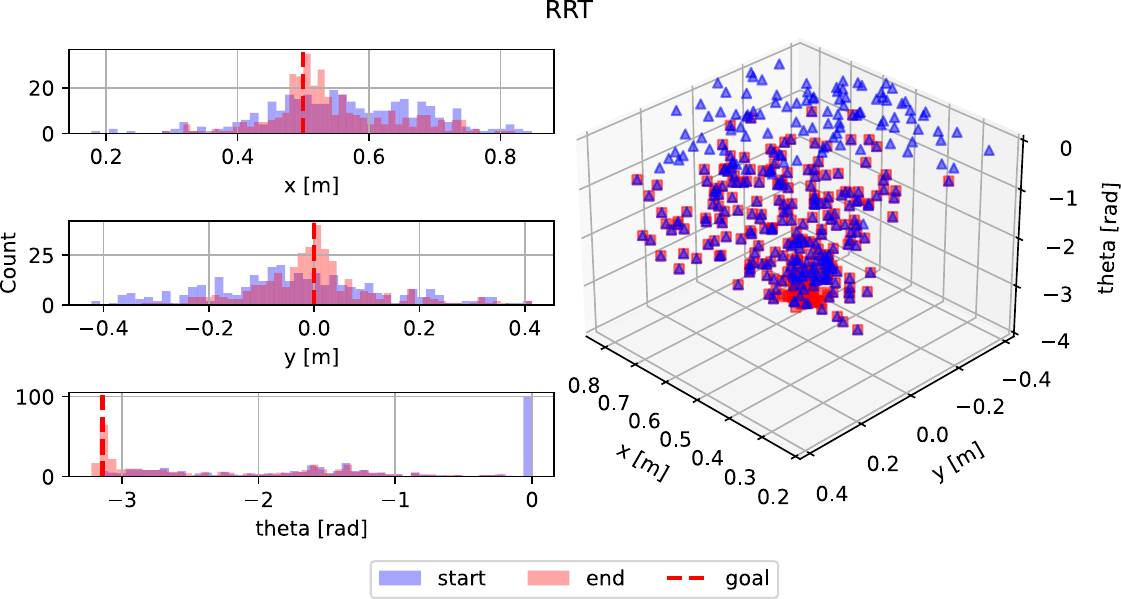}
    \caption{
    Distribution of reference trajectories generated by RRT.
    \textbf{Left}: Histograms showing the distribution of the start and end configurations of the object in the RRT-generated trajectories. The goal object configuration of $(0, 0, -\pi)$ is indicated.
    \textbf{Right}: 3D visualization of start and end object configurations of reference trajectories. While $x$ and $y$ coordinates are randomized,  $\theta$ is kept constant due to the cylinder's rotational symmetry.}
\label{fig:coverage-rrt}
\end{figure} 
Our pipeline is depicted in \fig{fig:data_pipeline}. We begin by sampling  $N_\text{test}=340$ reference trajectories using RRT in \cite{pang2022global} with $N_\text{test}$ different initial and goal states. RRT uses the nominal ``(\texttt{a})'' parameter in \tab{table:parameter_objects}. 
These yield state and feedforward control input sequences, $\{\left(\mathbf{x}^{\text{rrt}}_t, \mathbf{v}^{\text{rrt}}_t\right)_{1 \le t \le T}^n\}_{n=1}^{N_\text{test}}$ as illustrated in \fig{fig:coverage-rrt}. 
For each $n = 1,\dots, {N_\text{test}}$, we use $\{\left(\mathbf{x}^{\text{rrt}}_t, \mathbf{v}^{\text{rrt}}_t\right)_{1 \le t \le T}^n\}$ as a warm-start for computing trajectory-optimized plans $\left(\mathbf{x}^{\text{opt}}_t, \mathbf{v}^{\text{opt}}_t\right)_{1 \le t \le T}^n$. The horizons $T$ are determined by the RRT warm-start and vary across reference trajectories, but always lie in the range $T \in [4, 64]$ with a mean of $T = 26$.

For SP-TrajOPT, we solve \eqref{equation_single_TO} under the nominal ``(\texttt{a})'' parameter. For MP-TrajOPT, we solve \eq{equation_average_cost_particle} for all $N= 3$ parameters (note that this produces $N$ sequences of states; we select $\mathbf{x}^{\text{opt}}_t$ to be the one under the nominal shape). 
Both use the following cost function: 
\begin{align}
    % J(\{\mathbf{x}_{t}\}_{t=1}^T,\{\mathbf{v}_{t}\}_{t=0}^{T-1}) = 
    \sum_{t=1}^T \|\mathbf{x}_t - \mathbf{x}_T^{\mathrm{rrt}}\|_{\mathbf{Q}_t}^2 + \sum_{t=0}^{T-1} \|\mathbf{v}_t\|_{\mathbf{R}_t}^2
\end{align}
where $\mathbf{Q}_t = \text{diag}\left(10, 10, 10, 0.1, 0.1, 0.1, 0.1, 0.1, 0.1\right)$, $\mathbf{R}_t = \text{diag}\left(5, 5, 5, 5, 5, 5\right), \forall t \in [0, \ldots, T-1]$, $\mathbf{Q}_T=\text{diag}\left(1000, 1000, 1000, 0.1, 0.1, 0.1, 0.1, 0.1, 0.1\right)$.
We formulate trajectory optimization programs using \drake{}'s \texttt{MathematicalProgram} \cite{drake} and solve them using SNOPT \cite{gill2005snopt}. We optimize with dynamic smoothing parameter $\kappa = 10^5$.
% this leads to \emph{low smoothing} and less force at a distance for more realistic dynamics. 
We use $h=0.1$ seconds as the time step.

\subsection{Feedback Synthesis} For  LQR, we compute gains $\{\left(\mathbf{K}_t^\text{lqr}\right)^i\}_{i=1}^{N_\text{test}}$ from $\{\left(\mathbf{x}^{\text{opt}}_t, \mathbf{v}^{\text{opt}}_t\right)^i\}_{i=1}^{N_\text{test}}$.
% from TrajOpt via \eq{eq:feedback_controller_lqr}. 
% gains for KP-LQR are computed analogously, using the dynamics in \eqref{eq:bar_dynaimcs}. 
It means that, when the trajectories are supplied by SP- or MP-TrajOPT, gains are computed around the nominal parameter dynamics, (\texttt{a}) parameter in \tab{table:parameter_objects}. We use $h=0.1$ seconds as the time step, but use a smaller smoothing parameter, $\kappa = 800$, for smoother derivatives. We \revise{tune and set} $\mathbf{Q}_t = \text{diag}\left(10, 10, 10, 0.1, 0.1, 0.1, 0.1, 0.1, 0.1\right), 
\mathbf{R}_t = \text{diag}\left(5, 5, 5, 5, 5, 5\right), \forall t \in [0, \ldots, T-1]$, $ 
\mathbf{Q}_T=\text{diag}\left(1000, 1000, 1000, 0.1, 0.1, 0.1, 0.1, 0.1, 0.1\right)$ \revise{to achieve good performance across $N_\text{test}$ reference trajectories}.

\subsubsection{Execution in \drake}\label{sec:Controller Transfer}
While plans and gains are synthesized via smoothed quasi-dynamic dynamics, we evaluate performance in either (i) \drake{} or (ii) on real hardware. The planning/feedback synthesis phase returns a sequence of states, feedforward control inputs, and feedback gains at discrete knot points $\left(\left\{\mathbf{x}_t\right\}_{t=0}^{T}, \left\{\mathbf{v}_t\right\}_{t=0}^{T-1}, \left\{\mathbf{K}_t\right\}_{t=0}^{T-1}\right)$.

We adopt a higher control loop frequency in \drake{} simulations, which necessitates interpolation of these discrete-time quantities. To do so,  we convert the knot points into continuous time using First-Order Hold (FOH): states $\mathbf{x}^\text{FOH}(t)$, feedforward control input $\mathbf{v}^\text{FOH}(t)$, and feedback control input trajectories $\mathbf{K}^\text{FOH}(t)$. 
% using first-order hold: $\mathbf{x}^\text{FOH}(s)$, 
% \tpcomment{Consider converting the sim subscript to a superscript? Both rrt and opt in Fig 2. are superscripts. } 
% $\mathbf{v}^\text{FOH}(s)$. 
For $\mathbf{K}^\text{FOH}(t)$, we interpolate elements of $\mathbf{K}_t$ using FOH. 
% $\mathbf{K}^\text{FOH}: t \in \left[0, T-1\right] \rightarrow \mathbb{R}^{n_\mathrm{a} \times \left(n_\mathrm{u} + n_\mathrm{a}\right)}$.
At each time step in a \drake{} simulation, we measure the state $\mathbf{x}^\text{mea}(t)$ and rollout out the controller $\mathbf{u}(t) = \mathbf{v}^\text{FOH}(t) + \mathbf{K}^\text{FOH}(t) \left(\mathbf{x}^\text{mea}(t) - \mathbf{x}^\text{FOH}(t)\right)$. 
% $\mathbf{x}^\text{rollout}: t \in \left[0, T\right] \rightarrow \mathbb{R}^{n_\mathrm{u} + n_\mathrm{a}}$ 
% \tpcomment{So "real" is closed-loop rollout, and "sim" is First-order Hold? Can we call them "rollot/sim" and `FOH`? People often associate the word "real" with hardware.}
\subsubsection{Evaluation Metrics}
To evaluate the performance of controllers, we define the following metrics:
\begin{subequations}
\begin{flalign}
%     \delta_\text{terminal}:= \frac{1}{T}\sum_{t=0}^T d\left(
% \mathbf{x}^{\mathrm{u}}_{\text{rollout}, t}, \mathbf{x}^{\mathrm{u}}_{\text{reference}}{\text{drake}, t}
%     \right)\\
    \delta_\text{terminal}:= d\left(
\mathbf{x}^{\mathrm{u}, \text{mea}}_{T}, \mathbf{x}^{\mathrm{u}, \text{reference}}_{T}
    \right)
    \\
    \Delta_\text{terminal}:= \frac{1}{N_\text{test}} \sum_{i=1}^{N_\text{test}}
    \delta_\text{terminal}^i
    \\
    \eta^{\mathrm{B}}_\mathrm{A} := \frac{\Delta_\text{terminal}^\mathrm{B}}{\Delta_\text{terminal}^\mathrm{A}}
    \label{eq:eta}
    \end{flalign}
\end{subequations}
where 
% $\delta_\text{terminal}$ is the mean tracking error, 
$\delta_\text{terminal}$  is the tracking error at the terminal time step. The superscript $\mathrm{u}$ extracts the elements corresponding to $\mathbf{q}^\mathrm{u}$ 
and $\mathbf{x}^{\mathrm{u}, \text{reference}}_{T}$ is the reference trajectory the controller tries to track.
Through experiments, we consider $\mathbf{x}^{\mathrm{u}, \text{reference}}_{T} = \mathbf{x}^{\mathrm{u}, \text{rrt}}_{T}$.
$d\left(\cdot, \cdot\right)$ computes the Cartesian and angular displacements between two object poses.
% position and orientation given two configurations of the object. \tpcomment{Cartesian and angular displacements between two object poses}
Given $N_\text{test}$ reference trajectories 
% \hzcomment{plans instead of demonstrations, we are not writing in the context of imitation learning so the term demonstration does not make sense here}
computed from RRT, 
$\delta_\text{terminal}^i$ shows that it is the tracking error of the $i$-th result. 
$\Delta_\text{terminal}$ represents the average terminal tracking error over $N_\text{test}$ demonstrations. 
Given two ${\Delta_\text{terminal}^\mathrm{A}}$ and ${\Delta_\text{terminal}^\mathrm{B}}$ obtained from controller $\mathrm{A}$ and $\mathrm{B}$ respectively, relative cost,  $\eta^{\mathrm{B}}_\mathrm{A}$ represents how much the tracking errors are different between controller $\mathrm{A}$ and $\mathrm{B}$.
% \mscomment{clean this up}

\subsubsection{Hardware Setup}
We use two 7-DoF Kuka iiwa arms for the hardware experiments. The robots run a joint impedance controller.  We use a motion capture system to measure the state of the objects. 
% We execute our controller in the robots using the same execution in Section \ref{sec:Controller Transfer}.
% \mscomment{maybe reiterate how we do interpolation here as well} 

\subsection{Robustness Tests}
To evaluate the robustness of controllers, we consider the perturbations to initial states and the radius of the cylinder. 
% For both tests, we compute $\mathbf{x}^\text{opt}$, $\mathbf{v}^\text{opt}$, and $\mathbf{K}^\text{lqr}$  as explained in Section \ref{sec:data_generation_pipeline}, and we
% calculate $\mathbf{x}^\text{FOH}$,  $\mathbf{v}^\text{FOH}$, and $\mathbf{K}^\text{FOH}$
% as explained in Section \ref{sec:Controller Transfer}.

\subsubsection{Robustness to Initial Condition}\label{sec:robustness_initial_condition_explanation}
After we obtain $\mathbf{x}^\text{FOH}(t), \mathbf{v}^\text{FOH}(t), \mathbf{K}^\text{FOH}(t)$ (see Section \ref{sec:Controller Transfer}), 
we add perturbations $\left(\delta x, \delta y, \delta \theta\right)$ to the initial states of cylinder in \drake{} (i.e., $\textbf{x}^{\mathrm{u}, \text{mea}}(0) \leftarrow \textbf{x}^{\mathrm{u}, \text{mea}}(0) + [\delta x, \delta y, \delta \theta]^\top$)
% , $\textbf{x}^{\mathrm{u}, \text{mea}}$ is the elements of $\textbf{x}^{\text{mea}}$ corresponding to the object states
 and then we start rolling out the controller in  \drake{}.
% It means that at $t = 0$, we rollout $\mathbf{u}_t = \mathbf{v}_t + \mathbf{K}_t \left(\mathbf{x}_t - \mathbf{x}_0\right)$
We consider 50 points $\left(\delta x, \delta y, \delta \theta\right) \sim \mathrm{Uniform}[[-\delta x_0, + \delta x_0]  \times [-\delta y_0, + \delta y_0] \times [-\delta \theta_0, + \delta \theta_0] )$ per reference trajectory where $\delta x_0 = \delta y_0= \SI{0.025}{\meter}, \delta \theta_0 = \SI{5}{\degree}$.

\subsubsection{Robustness to Variation in Shape}
After we obtain $\mathbf{x}^\text{FOH}(t), \mathbf{v}^\text{FOH}(t), \mathbf{K}^\text{FOH}(t)$, we add perturbations $\delta r$
to the radius of cylinder, $r$, in \drake{} (i.e., $r \leftarrow r + \delta r$) and then we start rolling out the controller in  \drake{} with this updated $r$.
We consider 20 points $\delta r,  \sim \mathrm{Uniform}[[-\delta r_0, + \delta r_0])$ per reference trajectory where $\delta r_0 = \SI{0.01}{\meter}$.

\subsection{Results of SP-TrajOPT}\label{sec:result_sp_trajopt}
\begin{figure}[t]
    \centering
    \includegraphics[width=0.4\textwidth]{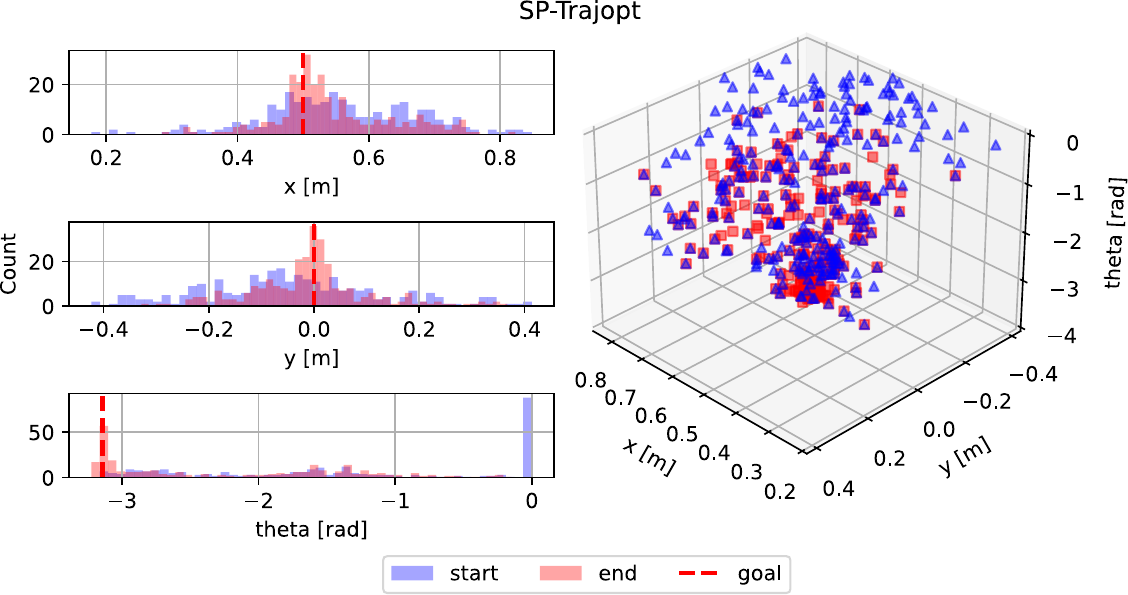}
    \caption{Reference trajectories on which SP-TrajOPT ran successfully.}
\label{fig:coverage-sp}
\end{figure}

% We first discuss the results of SP-TrajOPT.
\fig{fig:coverage-sp} shows the coverage of  SP-TrajOPT.  \fig{fig:coverage-rrt} and \fig{fig:coverage-sp} shows that SP-TrajOPT could successfully converge given reference trajectories by RRT.
The success rate of SP-TrajOPT given 340 reference trajectories is 82.8\%. 
We believe that this success rate is relatively high compared to other contact-implicit trajectory optimization frameworks. 
% To the best of the authors' knowledge, this is the highest success rate among contact-implicit trajectory optimization frameworks. 
We think this is because the contact complementarity constraints are implicitly imposed through $f_\kappa$ in our formulation.
% , enabling the optimizer to avoid evaluating KKT matrix with complementarity constraints. 
% This is because we use contact smoothing and SNOPT, which is one of the best off-the-shelf solvers.
We also observe that 
the average position and orientation errors of the cylinder are reduced, from \SI{0.18}{\meter} and \SI{71.9}{\degree} with RRT to \SI{0.14}{\meter} and \SI{55.0}{\degree} with SP-TrajOPT, respectively. Hence, our SP-TrajOPT could successfully generate more optimal trajectories.

% the average length of position and orientation of the cylinder over the reference trajectories with SNOPT success are \SI{0.18}{\meter} and \SI{71.9}{\degree} with RRT and \SI{0.14}{\meter} and \SI{55.0}{\degree} with SP-TrajOPT. Hence, our SP-TrajOPT could successfully generate more optimal trajectories.
% of the cylinder.

% Also, the success rate of SP-TrajOPT given 340 reference trajectories is 82.8 \%. 
% To the best of the authors' knowledge, this is the highest success rate among contact-implicit trajectory optimization frameworks. This is because we use contact smoothing and SNOPT, which is one of the best off-the-shelf solvers.

\subsection{Results of LQR with SP-TrajOPT}
\begin{figure}
     \begin{subfigure}[b]{0.24\textwidth}
         \centering
         \includegraphics[width=0.825\textwidth]{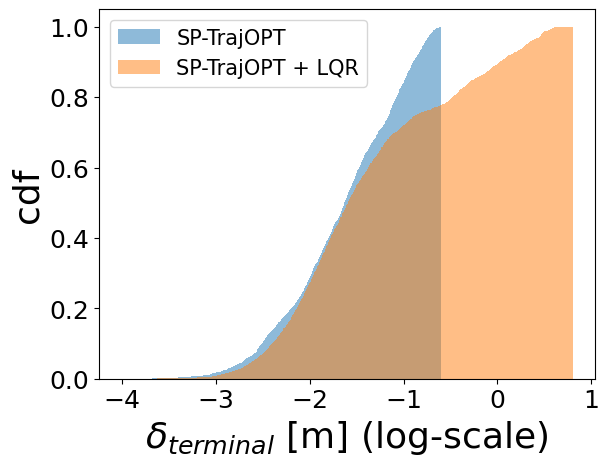}
         % {figure/dtrajopt_vs_rtrajopt_box_initial_state_noise_position.png}
         \caption{Terminal position error. $\Delta_\text{terminal}$ of SP-TrajOPT and SP-TrajOPT with LQR is \SI{0.046}{\meter} and \SI{0.30}{\meter}, respectively.}
\label{fig:sp_vs_splqr_initial_original_position}
     \end{subfigure}
     \hfill
     \begin{subfigure}[b]{0.24\textwidth}
         \centering
         \includegraphics[width=0.825\textwidth]{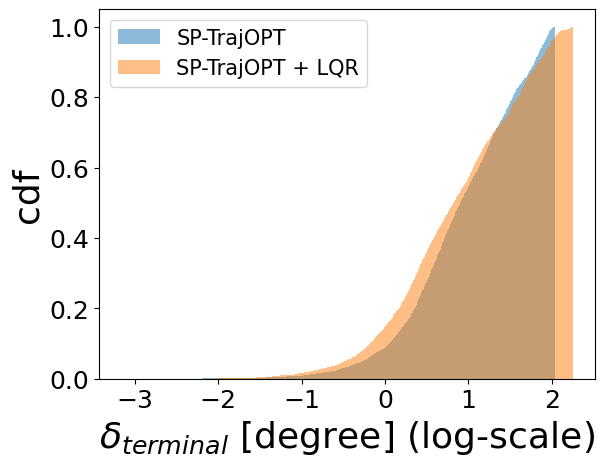}
         % {figure/dtrajopt_vs_rtrajopt_box_initial_state_noise_orientation.png}
         \caption{Terminal orientation error. $\Delta_\text{terminal}$ of SP-TrajOPT and SP-TrajOPT with LQR is \SI{20.9}{\degree} and \SI{17.1}{\degree}.}
         \label{fig:sp_vs_splqr_initial_original_orientation}
     \end{subfigure}
        \caption{We evaluate terminal pose tracking error $\delta_{\text{terminal}}$  of cylinder using SP-TrajOPT with / without LQR under perturbation of initial conditions.
        Note that cdf stands for cumulative distribution function.
        }
\label{fig:sp_vs_splqr_initial_original}
\end{figure}

\begin{figure}
     \begin{subfigure}[b]{0.24\textwidth}
         \centering
         \includegraphics[width=0.825\textwidth]{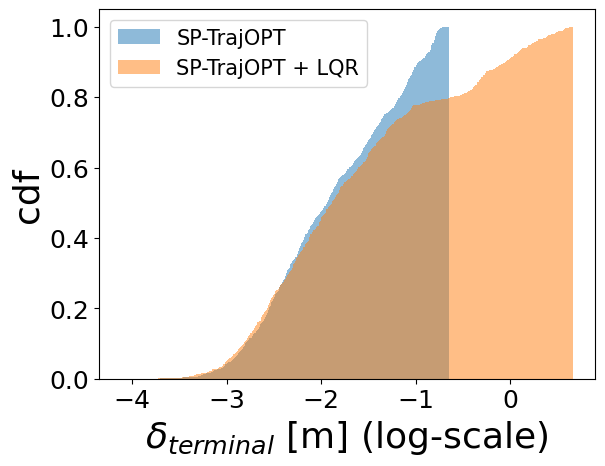}
         % {figure/dtrajopt_vs_rtrajopt_box_initial_state_noise_position.png}
         \caption{Terminal position error. $\Delta_\text{terminal}$ of SP-TrajOPT and SP-TrajOPT with LQR is \SI{0.034}{\meter} and \SI{0.21}{\meter}, respectively.}
\label{fig:sp_vs_splqr_shape_original_position}
     \end{subfigure}
     \hfill
     \begin{subfigure}[b]{0.24\textwidth}
         \centering
         \includegraphics[width=0.825\textwidth]{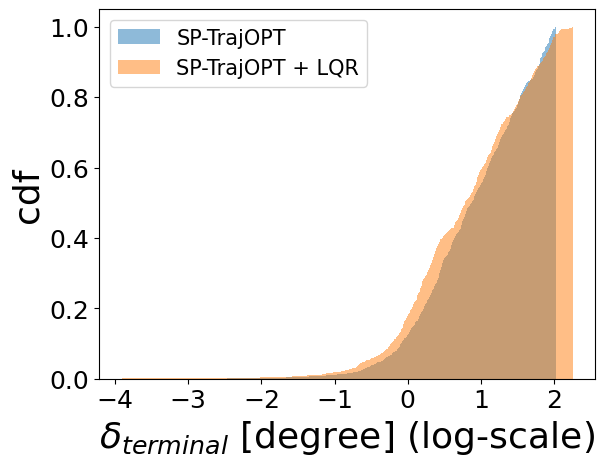}
         % {figure/dtrajopt_vs_rtrajopt_box_initial_state_noise_orientation.png}
         \caption{Terminal orientation error. $\Delta_\text{terminal}$ of SP-TrajOPT and SP-TrajOPT with LQR is \SI{19.7}{\degree} and \SI{20.2}{\degree}.}
         \label{fig:sp_vs_splqr_shape_original_orientation}
     \end{subfigure}
        \caption{We evaluate terminal pose tracking error $\delta_{\text{terminal}}$  of cylinder using SP-TrajOPT with / without LQR under variation in shape.}
\label{fig:sp_vs_splqr_shape_original}
\vskip -0.3 true in
\end{figure}

% We discuss the tracking error of LQR against SP-TrajOPT.

\subsubsection{Perturbations to Initial Conditions}

\fig{fig:sp_vs_splqr_initial_original} shows SP-TrajOPT and LQR across variations of initial conditions for the cylinder. In \fig{fig:sp_vs_splqr_initial_original_position}, SP-TrajOPT outperforms LQR. In \fig{fig:sp_vs_splqr_initial_original_orientation}, both controllers show similar performance while SP-TrajOPT with LQR shows slightly better $\Delta_\text{terminal}$.

\subsubsection{Perturbations to Shape}

The results of perturbations to the radius of the cylinder are shown in \fig{fig:sp_vs_splqr_shape_original}.
SP-TrajOPT outperforms SP-TrajOPT with LQR in both position and orientation tracking errors.

\subsection{Results of MP-TrajOPT}

\begin{figure}[t]
    \centering
    \includegraphics[width=0.38\textwidth]{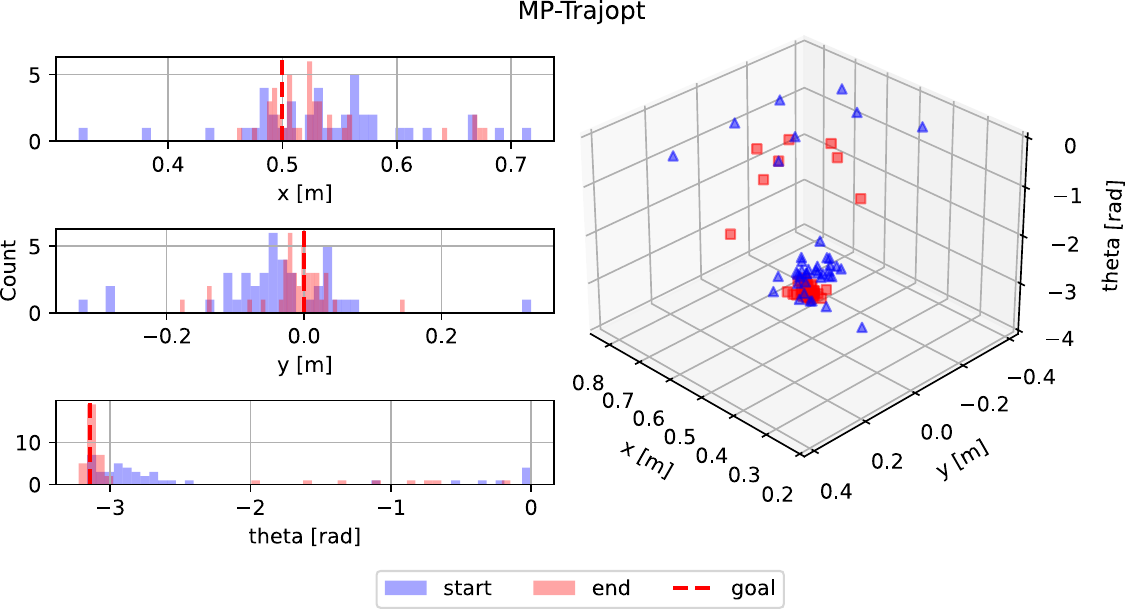}
    \caption{Reference trajectories on which MP-TrajOPT ran successfully.}
\label{fig:coverage-mp}
\end{figure} 

% To improve the performance of LQR, we believe that computing LQR gains on top of MP-TrajOPT-generated trajectories is important because MP-TrajOPT introduces some robustness against shape uncertainty. To verify this hypothesis, we study the effect of MP-TrajOPT in this section and Section \ref{sec:results_lqr_mp_trajopt}.

% We discuss the results of MP-TrajOPT. 
\fig{fig:coverage-mp} shows the coverage of MP-TrajOPT. Compared to the result of SP-TrajOPT in \fig{fig:coverage-sp}, we observe that MP-TrajOPT 
\revise{shows much smaller coverage.}
% could not show as good coverage as SP-TrajOPT could.
% 
The success rate of MP-TrajOPT 
% given 340 reference trajectories
is 10.6 \%, which is much lower than that of SP-TrajOPT. 
This is because MP-TrajOPT is much more complex than SP-TrajOPT, and thus, SNOPT might not be able to make any progress during optimization due to many reasons, such as poor scaling of the optimization problem. 
% and larger scale of the optimization problem.

\subsection{Results of LQR with MP-TrajOPT}\label{sec:results_lqr_mp_trajopt}

% This section discusses how MP-TrajOPT behaves. For comparison against SP-TrajOPT, we consider feasible trajectories that both SP- and MP-TrajOPT find.

\begin{figure}
     \begin{subfigure}[b]{0.24\textwidth}
         \centering
         \includegraphics[width=0.825\textwidth]{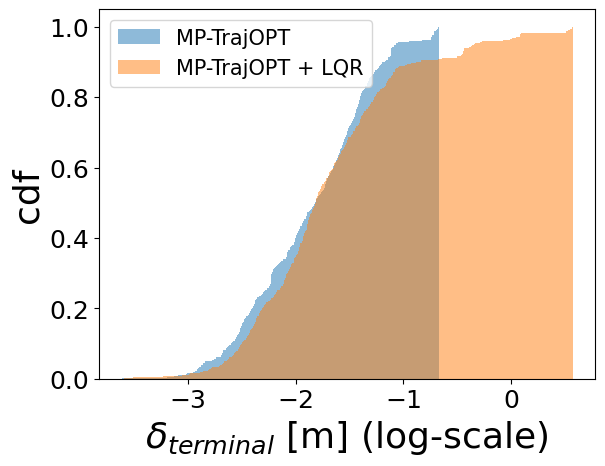}
         % {figure/dtrajopt_vs_rtrajopt_box_initial_state_noise_position.png}
         \caption{Terminal position error. $\Delta_\text{terminal}$ of MP-TrajOPT and MP-TrajOPT with LQR is \SI{0.027}{\meter} and \SI{0.10}{\meter}, respectively.}
\label{fig:mp_vs_mplqr_initial_original_position}
     \end{subfigure}
     \hfill
     \begin{subfigure}[b]{0.24\textwidth}
         \centering
         \includegraphics[width=0.825\textwidth]{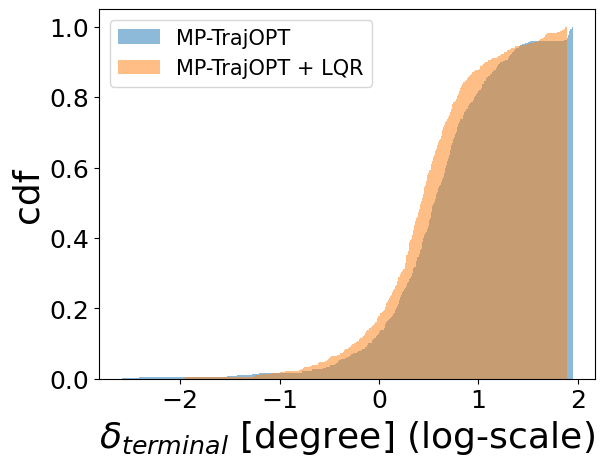}
         % {figure/dtrajopt_vs_rtrajopt_box_initial_state_noise_orientation.png}
         \caption{Terminal orientation error. $\Delta_\text{terminal}$ of MP-TrajOPT and MP-TrajOPT with LQR is \SI{8.9}{\degree} and \SI{5.2}{\degree}.}
         \label{fig:mp_vs_mplqr_initial_original_orientation}
     \end{subfigure}
        \caption{We evaluate terminal pose tracking error $\delta_{\text{terminal}}$  of cylinder using MP-TrajOPT with / without LQR under perturbation of initial conditions.}
\label{fig:mp_vs_mplqr_initial_original}
\vskip -0.3 true in
\end{figure}

\begin{figure}
     \begin{subfigure}[b]{0.24\textwidth}
         \centering
         \includegraphics[width=0.825\textwidth]{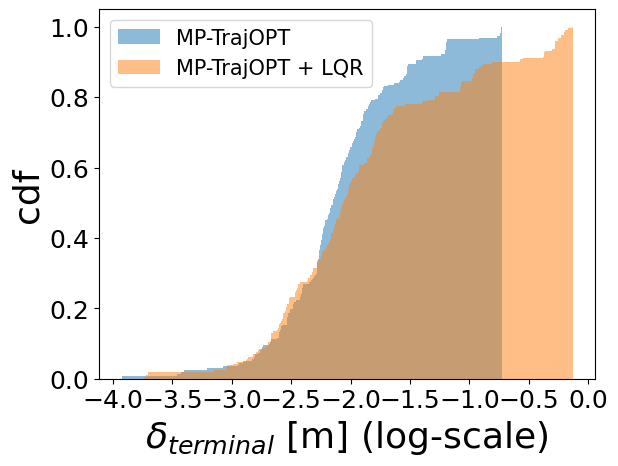}
         % {figure/dtrajopt_vs_rtrajopt_box_initial_state_noise_position.png}
         \caption{Terminal position error. $\Delta_\text{terminal}$ of MP-TrajOPT and MP-TrajOPT with LQR is \SI{0.019}{\meter} and \SI{0.059}{\meter}, respectively.}
\label{fig:mp_vs_mplqr_shape_original_position}
     \end{subfigure}
     \hfill
     \begin{subfigure}[b]{0.24\textwidth}
         \centering
         \includegraphics[width=0.825\textwidth]{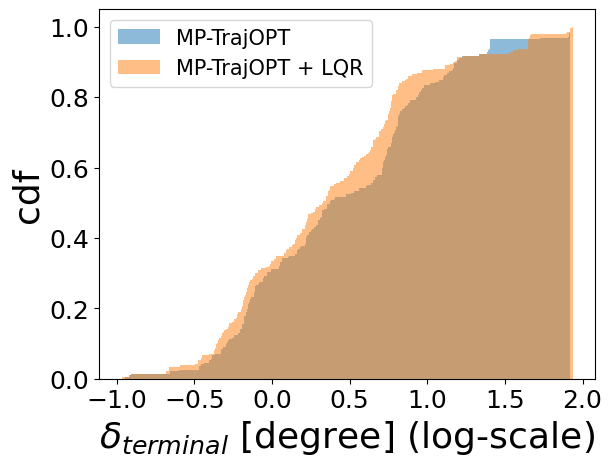}
         % {figure/dtrajopt_vs_rtrajopt_box_initial_state_noise_orientation.png}
         \caption{Terminal orientation error. $\Delta_\text{terminal}$ of MP-TrajOPT and MP-TrajOPT with LQR is \SI{8.8}{\degree} and \SI{6.8}{\degree}.}
         \label{fig:mp_vs_mplqr_shape_original_orientation}
     \end{subfigure}
        \caption{We evaluate terminal pose tracking error $\delta_{\text{terminal}}$  of cylinder using MP-TrajOPT with / without LQR under variation in shape.}
\label{fig:mp_vs_mplqr_shape_original}
\end{figure}

% We discuss the tracking error of LQR on top of trajectories generated from MP-TrajOPT. 

\subsubsection{Perturbations to Initial Conditions}\label{sec:perturabation_initial_condition}
The results of perturbations to initial conditions are shown in \fig{fig:mp_vs_mplqr_initial_original}.
While MP-TrajOPT outperforms MP-TrajOPT with LQR in \fig{fig:mp_vs_mplqr_initial_original_position}, the relative cost in \fig{fig:mp_vs_mplqr_initial_original_position}, $\eta^{{\text{MP-TrajOPT + LQR}}}_\text{MP-TrajOPT}  = 3.7$.  
Since $\eta$ in \fig{fig:sp_vs_splqr_initial_original_position} is $\eta^{\text{SP-TrajOPT + LQR}}_{\text{SP-TrajOPT}} = 6.5$, we argue that MP-TrajOPT improves the LQR performance. 
For the orientation tracking error, \fig{fig:mp_vs_mplqr_initial_original_orientation} shows that MP-TrajOPT with LQR outperforms MP-TrajOPT while in \fig{fig:sp_vs_splqr_initial_original_orientation}, SP-TrajOPT outperforms SP-TrajOPT with LQR. Therefore, we observe that MP-TrajOPT introduces some robustness, resulting in improved LQR performance. Also, $\eta$ in \fig{fig:mp_vs_mplqr_initial_original_orientation} is $\eta^{\text{MP-TrajOPT + LQR}}_{\text{MP-TrajOPT}} = 0.58$ while $\eta$ in \fig{fig:sp_vs_splqr_initial_original_orientation} is $\eta^{\text{SP-TrajOPT + LQR}}_{\text{SP-TrajOPT}} = 0.82$. 
Although our MP-TrajOPT is not designed to be robust against variation in initial conditions, the result suggests that the inherent robustness against parametric uncertainty of the model contributes to robustness under variation in initial conditions. 
% \revise{However, the above observations are only valid when MP-TrajOPT is solved.}
% Although MP-TrajOPT still better results in position, the position tracking error between MP-TrajOPT and SP-TrajOPT is smaller than the case considering SP-TrajOPT. MP-TrajOPT helps LQR stabilize the system.

\subsubsection{Perturbations to Shape}

The results are shown in \fig{fig:mp_vs_mplqr_shape_original}. We observe that the performance of LQR improves. The relative cost in \fig{fig:mp_vs_mplqr_shape_original_position} and \fig{fig:sp_vs_splqr_shape_original_position} is $\eta^{{\text{MP-TrajOPT + LQR}}}_\text{MP-TrajOPT} = 3.1$ and $\eta^{{\text{SP-TrajOPT + LQR}}}_\text{SP-TrajOPT} = 6.2$, respectively. Similarly, the relative cost in \fig{fig:mp_vs_mplqr_shape_original_orientation} and \fig{fig:sp_vs_splqr_shape_original_orientation} is $\eta^{{\text{MP-TrajOPT + LQR}}}_\text{MP-TrajOPT} = 0.77$ and $\eta^{{\text{SP-TrajOPT + LQR}}}_\text{SP-TrajOPT} = 1.01$, respectively. 
Therefore, we observe that MP-TrajOPT and LQR work synergistically.
\revise{However, these improvements can be observed only when MP-TrajOPT is solved, which is often difficult. 
% Ensuring reliable solutions to MP-TrajOPT remains an interesting area for future work.
}

\subsection{Hardware Experiments}
% We demonstrate our controllers in hardware experiments.
We implement two controllers, open-loop \revise{using SP-TrajOPT} and LQR controllers, and evaluate their tracking performance in hardware experiments. As shown in \fig{fig:whole-body}, we observe that LQR could track the specific reference trajectory with perturbations to the initial states.

\section{A Closer Look at LQR}\label{sec:closer_look}

\subsection{Results of Different Smoothing}\label{sec:different_smoothing_result}

\begin{figure}[t]
     \begin{subfigure}[t]{0.5\textwidth}
         \centering
         \includegraphics[width=0.6\textwidth]{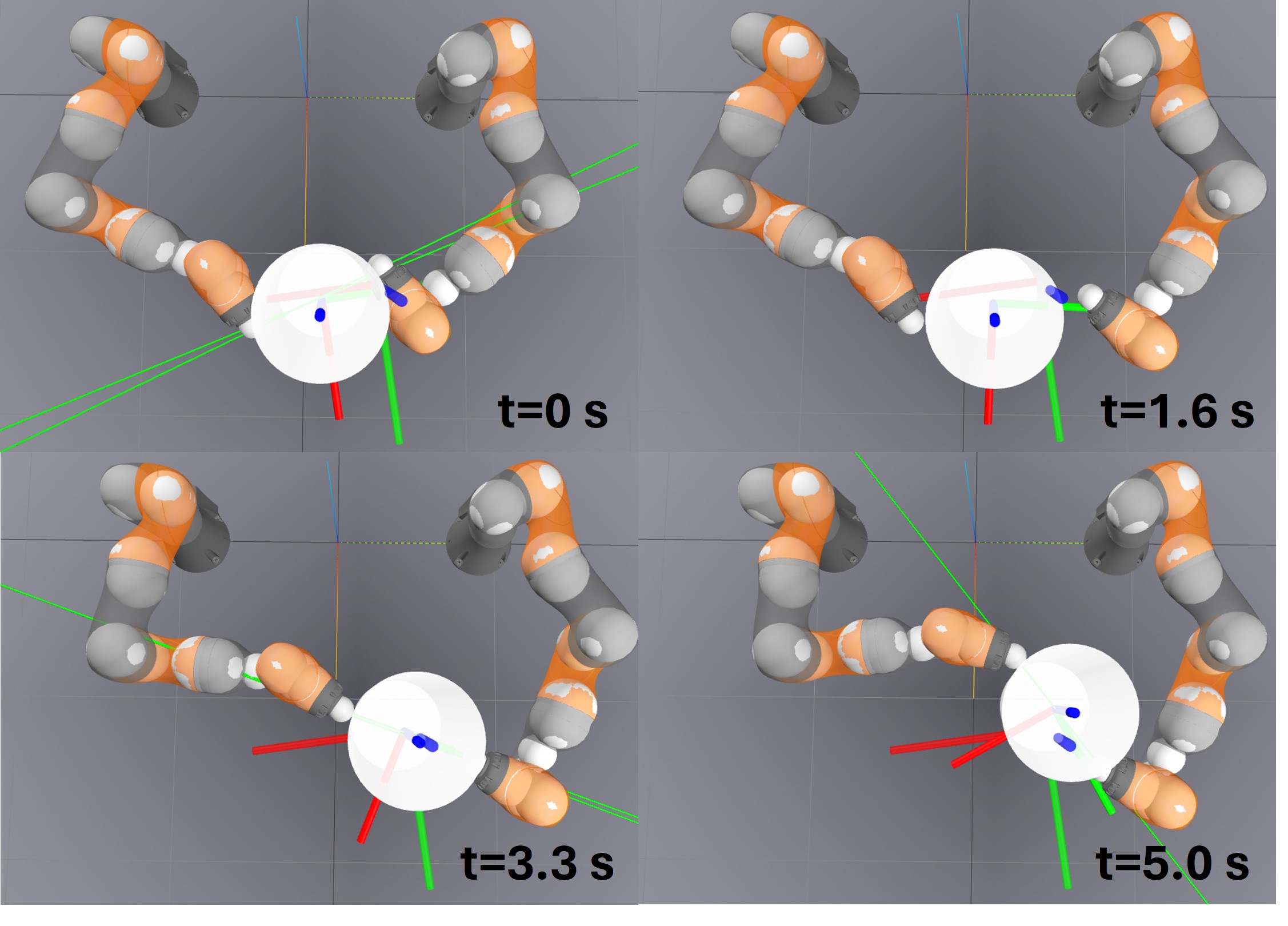}
         \vspace{-0.5em}
         \caption{Simulation result of LQR with cylinder with $\kappa=160$. }
\label{fig:kappa_160}
     \end{subfigure}
     % \hfill
     % \\
     \begin{subfigure}[t]{0.5\textwidth}
         \centering
         \includegraphics[width=0.6\textwidth]{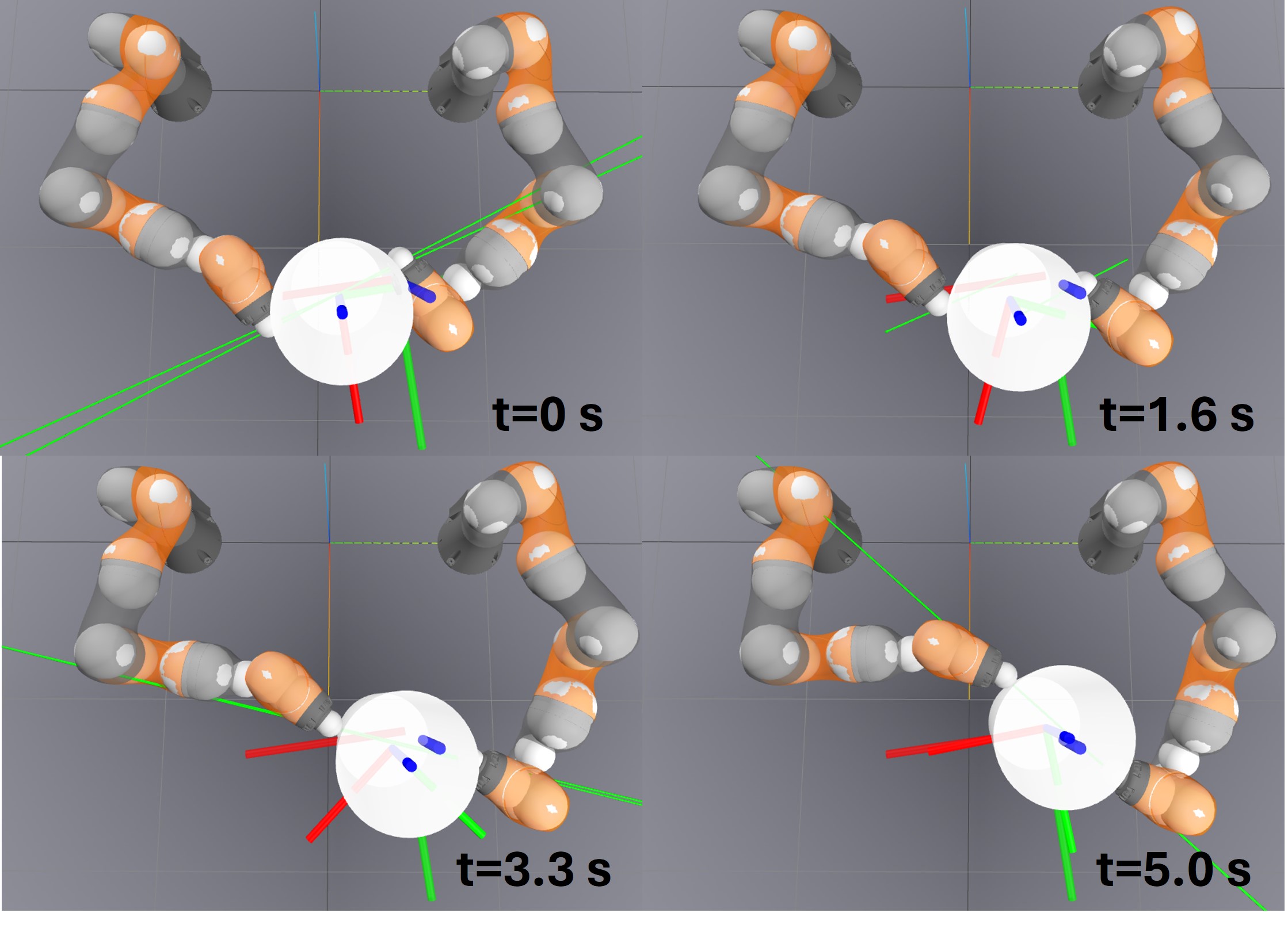}
         \vspace{-0.5em}
         \caption{Simulation result of LQR with cylinder with $\kappa=800$.}
         \label{fig:kappa_800}
     \end{subfigure}
        \caption{Snapshots along a trajectory simulated in \drake. We implement LQR with the cylinder with different smoothing parameters $\kappa$ for the same reference trajectory with the same initial condition perturbations. The snapshots show the cylinder frame and the desired frame at the terminal time step. 
        Green lines represent contact forces. 
        At $t=1.6$ s,  the right arm in \fig{fig:kappa_160} loses the contact and causes \textit{pulling} motion while the right arm in \fig{fig:kappa_800} maintains the contact. As a result, LQR in \fig{fig:kappa_160} fails to track the reference trajectory while LQR in \fig{fig:kappa_800} could successfully track it.}
        \label{fig:kappa}
\end{figure}
\begin{figure}[t]
     \begin{subfigure}[b]{0.2\textwidth}
         \centering
         \includegraphics[width=0.605\textwidth]{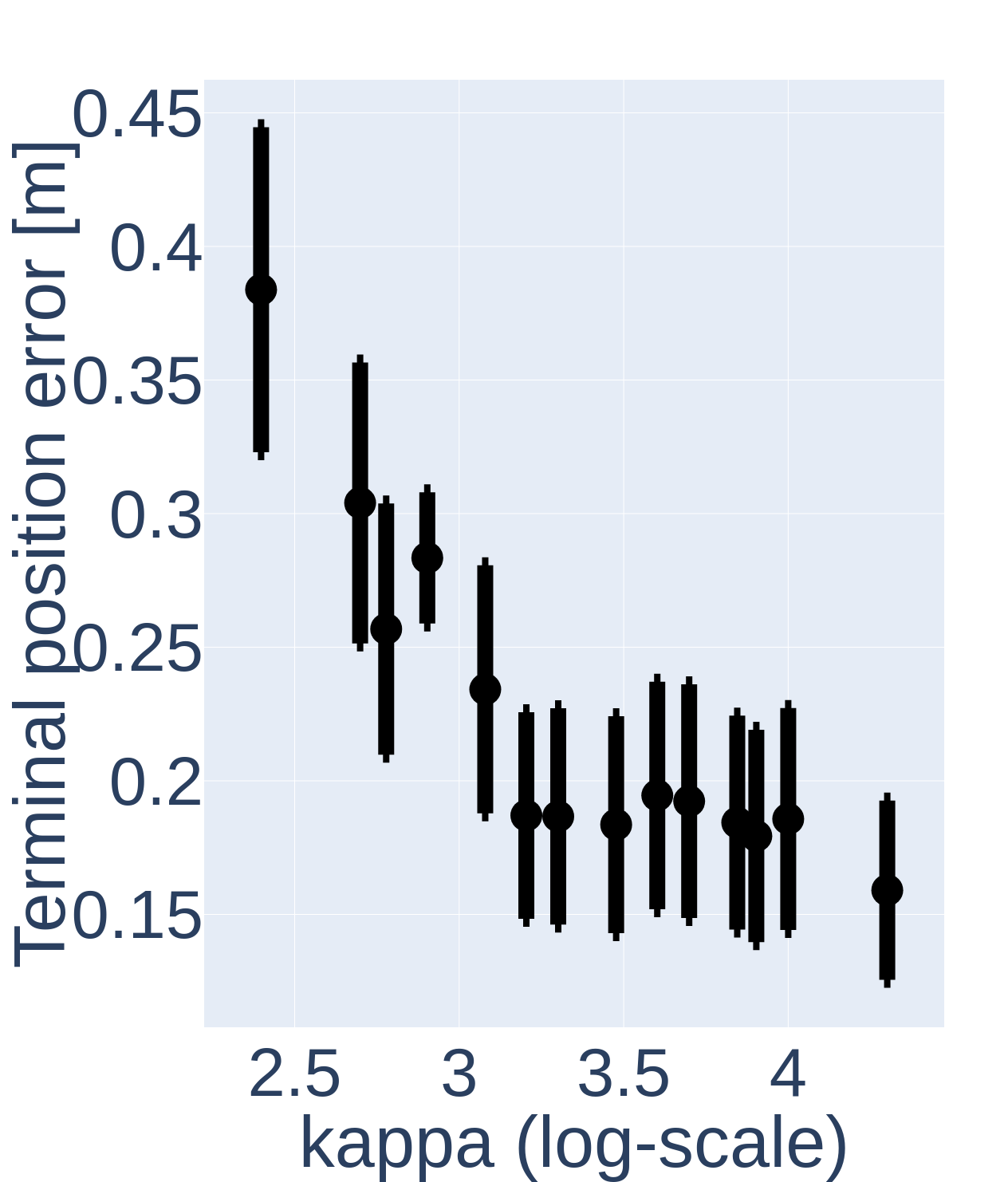}
         \caption{Terminal position error}
\label{fig:terminal_position_error_kappa_cylinder}
     \end{subfigure}
     \hfill
     \begin{subfigure}[b]{0.2\textwidth}
         \centering
         \includegraphics[width=0.605\textwidth]{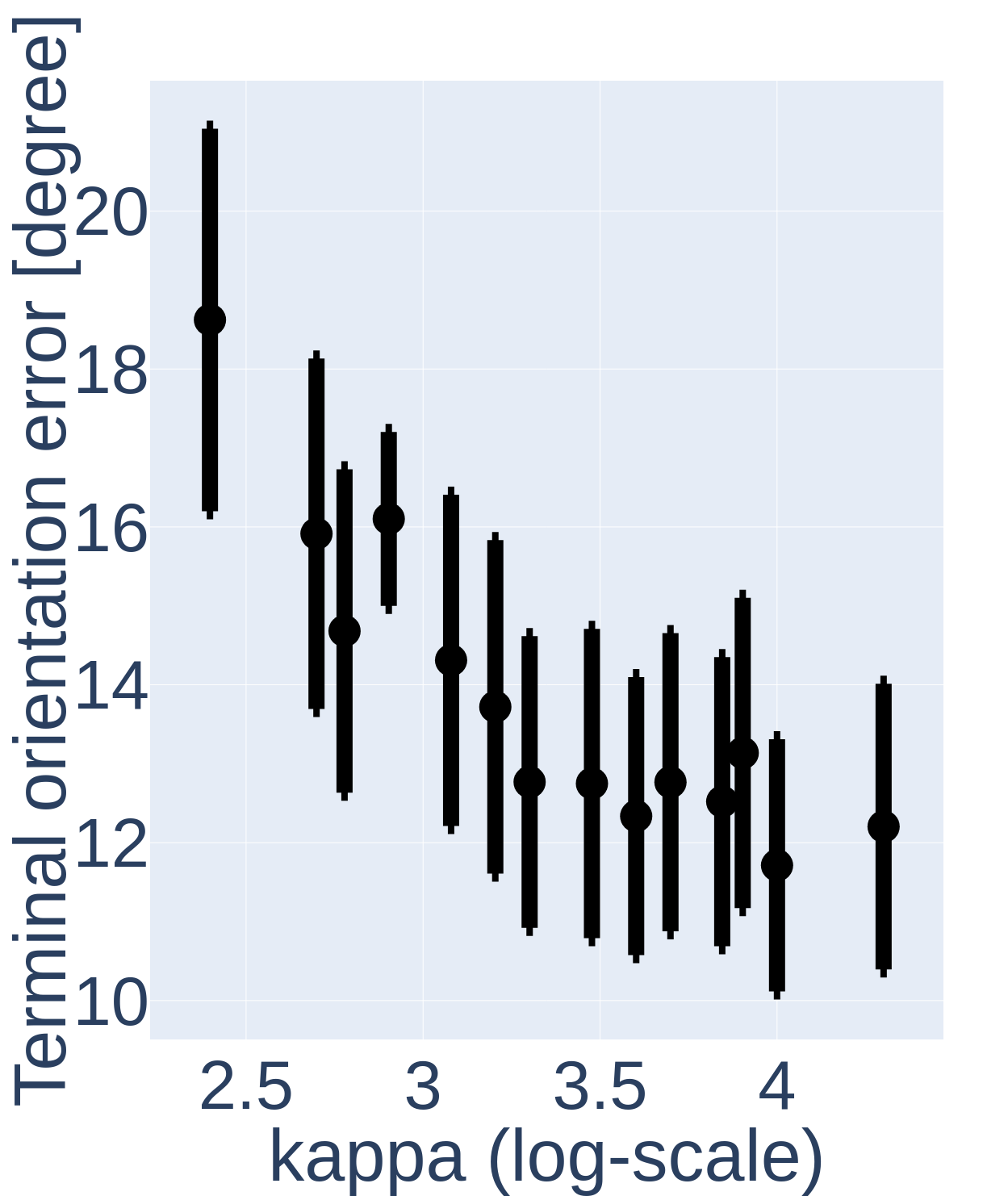}
         \caption{Terminal orientation error}
         \label{fig:terminal_orientation_error_kappa_cylinder}
     \end{subfigure}
        \caption{Terminal pose tracking error $\Delta_{\text{terminal}}$ with different smoothing values $\kappa$ using LQR with perturbations of initial conditions. In the figures, we show the bar indicating the mean and $95 \%$ confidence interval. }
        \label{fig:terminal_pose_error_kappa_cylinder}
\vspace{-1.5em}
\end{figure}

We discuss the relation between the behavior of LQR and the smoothing parameter $\kappa$. Here we have two results using LQR with $\kappa=160$ (i.e., more smoothing) and $\kappa=800$ (i.e., less smoothing) for the same reference trajectory with the same perturbations of initial conditions used in Section \ref{sec:robustness_initial_condition_explanation}. 
\fig{fig:kappa} shows the snapshots along the trajectory during the simulation in \drake.

% \textbf{Discussion.}
With $\kappa=160$, LQR makes the robot try to pull the cylinder even with no contact at $t = 1.6$ s in \fig{fig:kappa_160} due to large contact smoothing. 
% as we discuss in Section \ref{sec:badlinearization}. 
% This makes sense since \revise{connect with Terry's explanation?}
% 
% \revise{(we want to refer the figure which shows the gradient with smoothing?)}
% 
Through the experiments, we observe that a common rule of thumb for designing LQR with good tracking performance is to choose a large $\kappa$ - if $\kappa$ is too small, the controller will cause undesired pushing and \textit{pulling} motion of the robot 
% and if $\kappa$ is too large, the controller is not reactive and it does not do anything, 
as shown in \fig{fig:terminal_pose_error_kappa_cylinder}. 
% We analyze this undesired pulling motion in Section \ref{sec:badlinearization}.

\subsection{Fundamental Shortcomings of Linearization}\label{sec:badlinearization}
\begin{figure}[t]
    \centering
    \includegraphics[width=0.405\textwidth]{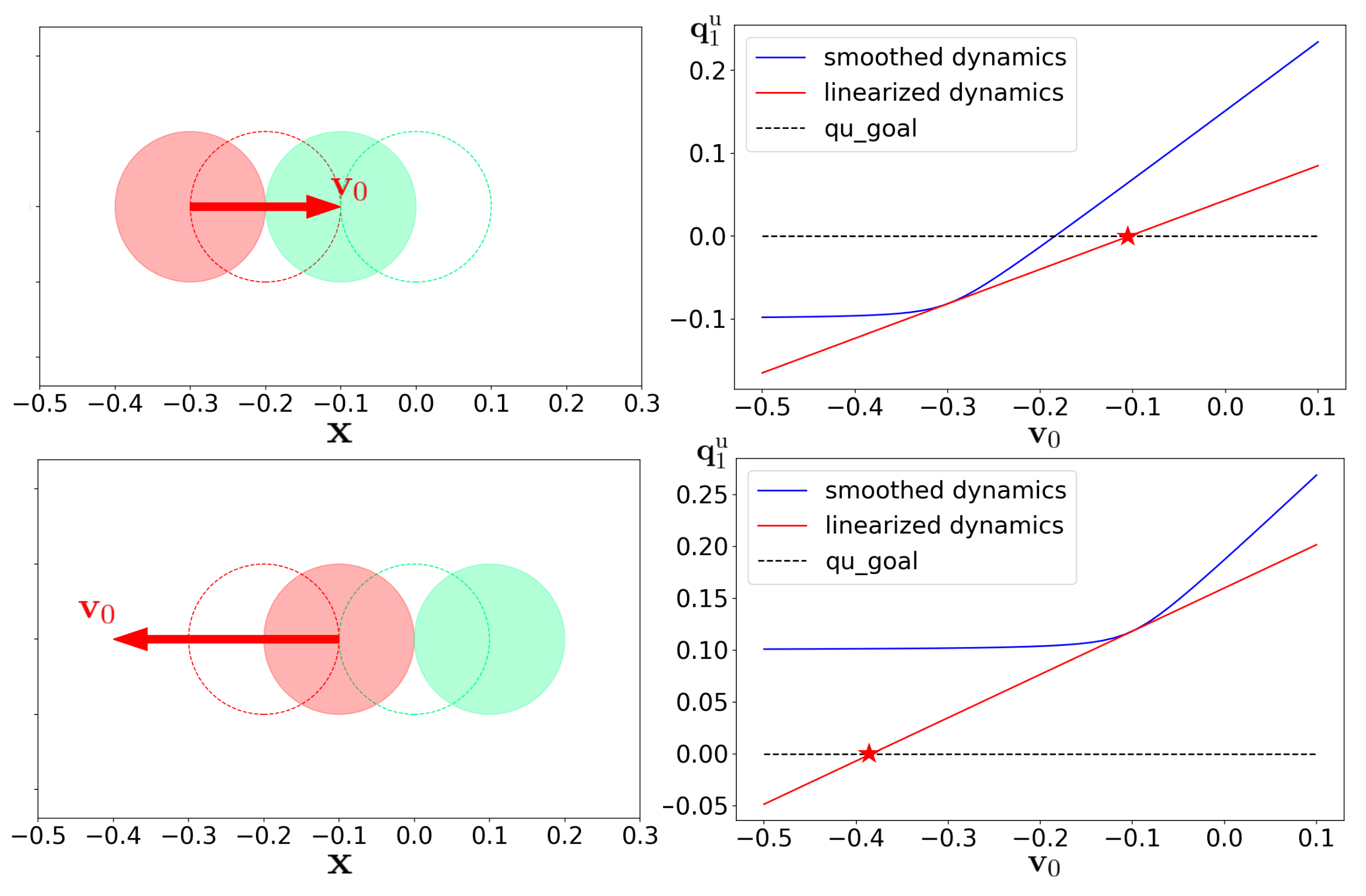}
    \caption{Illustration of how LQR behaves in a 1D ball-pushing environment, where the red actuated ball tries to push the green unactuated ball using control inputs derived from LQR. The dotted balls in the left column denote the nominal state where linearization is done, and the solid balls illustrate the current state. The right columns illustrate the smoothed dynamics, linearized dynamics, and the optimal LQR solution according to linearization. The top row and bottom row illustrate two cases of disturbances, where the top row can be stabilized by pushing, but the bottom row requires pulling.}
\label{fig:unilateral}
\vspace{-1.5em}
\end{figure}
We here analyze why LQR does not work well on top of SP-TrajOPT-generated trajectories.  
One limitation of LQR is that the linearization does not sufficiently capture the \emph{unilateral} nature of contact. To illustrate this, we consider a 1-step LQR problem in \eq{eq:lqronestep},
\begin{subequations}
\begin{flalign} \min_{\boldsymbol{\Delta} \mathbf{v}_{0}}
&\; \left\|\boldsymbol{\Delta} \mathbf{x}_{1} \right\|_{\mathbf{P}_1}^2
\\
\text{s.t.} &\; \boldsymbol{\Delta}\mathbf{x}_{1} = \mathbf{A}_{0}\boldsymbol{\Delta}\mathbf{x}_{0} + \mathbf{B}_{0}\boldsymbol{\Delta}\mathbf{v}_{0},\\
&\; \boldsymbol{\Delta}\mathbf{x}_{0} = \hat{\mathbf{x}} -\mathbf{x}_{0}.
\end{flalign}
\label{eq:lqronestep}
\end{subequations}
\noindent Consider a simple 1D block-pushing system (\fig{fig:unilateral}), where an actuated block is trying to push an unactuated block into a desired location. When we visualize the linearized dynamics in \fig{fig:unilateral}, we can observe that in cases where the red ball must \emph{push} to stabilize, this 1-step LQR takes a step in the right direction although overshooting occurs (\fig{fig:unilateral}, top row). If we ask LQR to recover from overshooting (\fig{fig:unilateral}, bottom row), however, a linearized model predicts that it can pull the object and take a step towards the opposite direction. Yet, due to the directional nature of contact, this has no effect on the unactuated ball. This mismatch between the linearized model and the true model leads to limitations of the LQR controller.
We observe this \textit{pulling} motion in \fig{fig:kappa} in Section \ref{sec:different_smoothing_result}.

The symmetric nature of local linear models is fundamentally at odds with the unilateralness of contact. We address this challenge in \cite{suh2025ctr}, where we build a \emph{trust region} in which the linearization is locally consistent with contact dynamics.

\section{Conclusion}\label{sec:discussion}
Is linear feedback on smoothed dynamics sufficient for stabilizing contact-rich plans? 
Our analysis and experiments suggest that designing LQR for contact-rich plans does not work well in general.  However, we observe that MP-TrajOPT enables LQR to improve its performance when MP-TrajOPT is solved, although MP-TrajOPT often fails to converge.
% designing LQR for robustified contact-rich plans works. 
% \tpcomment{ I don't think robustifition works well at all...? Can we tone down this statement? In addition to the low success rate, if we look at Fig 6 and 7, LQR is marginally better in orientation but a lot worse in position.}
Through this paper, we first present how contact smoothing technique can be used for designing trajectory optimization baselines and LQR. Then, we extensively conduct various experiments of LQR under different uncertainties. 
We hope that our analysis provides readers with insights into
design of planners and controllers for contact-rich manipulation
using differential simulators based on contact smoothing.

\section*{Acknowledgement}
The authors would like to thank Hongkai Dai for insightful discussions on nonlinear optimization.

\bibliographystyle{IEEEtran}
\bibliography{main.bib}
% \newpage
% \appendix
% \input{app_further_technical_discussion}
% \input{app_experiment_details}
% \input{app_additional_plots}
% \input{app_quasistatic_dynamics}

\end{document}